\documentclass[%
  reprint,           
  onecolumn,         
  nofootinbib,       
  superscriptaddress,
  amsmath,amssymb,   
  aps,
  authoryear,
  showkeys
]{revtex4-2}

\setcitestyle{authoryear,round}  

\usepackage[T1]{fontenc}

\usepackage{amssymb}
\usepackage{amsmath}

\usepackage[utf8]{inputenc}
\usepackage[T1]{fontenc}

\usepackage{algorithm}
\usepackage{algorithmicx}
\usepackage{algpseudocode}
\usepackage{color}
\usepackage{xcolor}
\usepackage{microtype}
\usepackage{hyperref}

\usepackage{booktabs}
\usepackage{graphicx}
\DeclareGraphicsExtensions{.pdf,.jpeg,.jpg,.png}
\graphicspath{{./figs/}}

\hypersetup{
  colorlinks=true,
  linkcolor=blue,
  citecolor=blue,
  urlcolor=blue
}

\begin{document}
\title{A Ranking-Based Optimization Algorithm for the Vehicle Relocation Problem in Car Sharing Services}

\thanks{%
This is the accepted manuscript of the article: 
P. Szwed, P. Skrzynski and J. Wąs: A ranking-based optimization algorithm for the vehicle relocation problem in car sharing services,  \textit{Transportation Research Part C: Emerging Technologies}, vol. 182, pp. 105421, 2026, ISSN 0968-090X , DOI \url{https://doi.org/10.1016/j.trc.2025.105421} © 2025 Elsevier Ltd. Licensed under CC BY-NC-ND 4.0.
}

\author{Piotr Szwed} 
\email{pszwed@agh.edu.pl}
\author{Paweł Skrzynski} 
\email{skrzynia@agh.edu.pl}
\author{Jarosław Wąs} 
\email{jaroslaw.was@agh.edu.pl}
\affiliation{AGH University of Krakow, Faculty of Electrical Engineering, Automatics, IT and Biomedical Engineering, al. Mickiewicza 30, 30-059 Kraków, Poland }

\begin{abstract}
The paper addresses the Vehicle Relocation Problem in free-floating car-sharing services by presenting a solution focused on strategies for repositioning vehicles and transferring personnel with the use of scooters. Our method begins by dividing the service area into zones that group regions with similar temporal patterns of vehicle presence and service demand, allowing the application of discrete optimization methods. In the next stage, we propose a fast ranking-based algorithm that makes its decisions on the basis of the number of cars available in each zone, the projected probability density of demand, and estimated trip durations. The experiments were carried out on the basis of real-world data originating from a major car-sharing service operator in Poland.
The results of this algorithm are evaluated against scenarios without optimization that constitute a baseline and compared with the results of an exact algorithm to solve the Mixed Integer Programming (MIP) model. As performance metrics, the total travel time was used.
Under identical conditions (number of vehicles, staff, and demand distribution), the average improvements with respect to the baseline of our algorithm and MIP solver were equal to 8.44\% and 19.6\% correspondingly. However, it should be noted that the MIP model also mimicked decisions on trip selection, which are excluded by current services business rules.  
The analysis of results suggests that, depending on the size of the workforce, the application of the proposed solution allows for improving performance metrics by roughly 3\%-10\%.

\end{abstract}


\maketitle

\section{Introduction}
\label{sec:intro}

In recent years, car-sharing systems have become an integral part of urban mobility solutions, offering an alternative to traditional private car ownership and public transportation. Mobility-on-Demand (MoD) services, including free-floating car-sharing systems that are gaining popularity worldwide, 
provide users with flexible short-term vehicle rentals, enhancing urban transportation efficiency and sustainability \citep{Baumgarte2022, Castagna2021}.


Car-sharing services may either compete with or complement transportation methods such as public transit and private vehicle ownership. Nonetheless, customer satisfaction and the operational success of these services hinge largely on the accessibility of vehicles during peak demand periods and at sought-after locations. In this context, the ongoing challenge is addressing the Vehicle Relocation Problem (VReP) and finding effective solutions for it \citep{ILLGEN2019193,Borghetti2022}

The stochastic and asymmetric nature of demand, coupled with variable travel times, results in an inevitable imbalance between supply and demand in specific regions for car-sharing services. To improve system performance, operators must immediately make the best vehicle allocation and relocation decisions in real time. 

This paper examines a staff-based operational framework for a VReP \citep{Borghetti2022, weikl2015, weikl20152}. In this system, personnel use electric scooters to travel to spots where the number of vehicles exceeds demand, then transfer vehicles to areas with higher demand. Operational decisions involve prompt scheduling of staff tasks. For both scooter transits and car relocation, shorter trips are prioritized to minimize vehicle relocation costs and ensure efficient use of staff resources.

The proposed solution was designed for integration within a car-sharing software platform operating in Poland. This platform supports more than 6,000 vehicles and has a user base that exceeds 500,000 registered members, with approximately 20\% of them being active users.
The analysis described in this paper is based on real-world data spanning from 2019 to 2024 and including detailed user activity data, such as logins, vehicle searches, reservations, and completed rentals.  As a case study, we focus on Kraków, a city of about 800,000 permanent inhabitants in southern Poland.

The relocation algorithm serves as an optimization technique designed to equalize disparities between demand and supply across various locations. Applying it necessitates decisions regarding the dimensions and forms of the locations under consideration.
In the current platform, data regarding available vehicles and user interactions are gathered from hexagonal zones with a side of 500 meters. These are visualized as part of the user interface. Nonetheless, these regions, each covering roughly 0.65 km², are too small to accurately forecast the demand for vehicles and their availability. It can be noted that the predictive accuracy diminishes with reduced zone sizes as a result of heightened variance and insufficient data density. Furthermore, these hexagonal regions are too numerous (specifically 365 in the case of Kraków) to effectively utilize classical optimization techniques with Mixed Integer Programming (MIP). In the scholarly literature, MIP approaches are often applied to systems with 15--20 discrete locations; nevertheless, computation times can range from several minutes to a few hours. This makes MIP impractical for real-time applications servicing larger cities.

\subsection{Goal}
Our main objective was to create a relocation algorithm that could be implemented in real-life free-floating car-sharing (FFCS) operations. When devising the solution, we focused on three key considerations: time efficiency, the limitations of demand modeling in the case of FFCS, and the specificity of local conditions. 

First, we assumed that the decision must be formulated and delivered to a relocation operator via a mobile application relatively quickly, within a maximum span of several seconds. We prioritize achieving a rapid response time over the complexity of the algorithm and the possibility of marginally improving efficiency.

Second, we planned to design a solution for a real-world FFCS service that does not support reservations (except for cargo vehicles) and does not rely on stations or dedicated parking spots. As a result, relocation decisions must be based entirely on forecasts of client demand and the predicted distribution of parked vehicles. Events in the car-sharing system, such as client interactions and vehicle parking, occur in a space that can be considered continuous. Therefore, these events can either be modeled using a fully continuous probability density function, or the time and space dimensions can be discretized into a grid of geographical zones and time intervals. This allows events to be represented as counts within the corresponding bins. 

As further discussed in Section~\ref{sec:demand-estimation}, prediction algorithms face limitations in efficiency when applied to fine-grained time-space grids. Typically, excellent results can be achieved either for large areas over short time horizons (e.g., predicting demand across an entire city within the next hour), or for small areas over longer intervals (e.g., predicting demand at a specific station over a week). In order to balance spatial granularity with prediction efficiency, we adopted a trade-off approach. Specifically, we decided to divide the service area into a few dozen larger zones, for which client demand and vehicle supply can be predicted more effectively.

The third factor influencing our solution was the local context. Kraków, like other major cities in Poland, faces a severe shortage of parking spaces. Although the official population of Kraków is around 800,000, the actual number of inhabitants is estimated to exceed 1 million. The city has approximately 650,000 registered vehicles, though the true number is likely higher due to cars owned by temporary residents. As a result, finding a parking spot at a specific location and time is highly uncertain. Even if a relocation algorithm suggests an ideal drop-off point, the driver may still need to search for parking across several nearby streets in practice. 

Since electric vehicles make up a small percentage of most car-sharing fleets in Polish cities—specifically, less than 5\% for Traficar—we opted to create a versatile algorithm that is not tailored to any specific type of vehicle. Nevertheless, anticipating emerging requirements, we also addressed the issue of adaptability. The algorithm can be easily extended to prioritize the relocation of electric vehicles, thereby reducing urban emissions and enhancing the synergy of operational strategies with sustainability objectives.







\subsection{Solution concept}

For optimization purposes, we assumed that the service area will be divided into a few dozen coherent zones, each characterized by short road distance between internal locations and similar observed patterns of demand and user activity. 
In particular, the proximity of locations within a single zone is an important factor, as it is related to the short walking distance expected by customers and increases the likelihood that cruising for parking while relocating the vehicle will end up in the desired spot.

The relocation process comprises two stages:

\begin{itemize}
 \item \textbf{Stage 1:} Inter-zone relocation of vehicles to match predicted demand distribution,
 \item \textbf{Stage 2:} Intra-zone vehicle placement based on user convenience and predicted micro-location demand.
\end{itemize}

This method's benefit lies in its use of a limited number of zones, facilitating improved estimation of travel demand between these regions and enabling the application of various optimization algorithms, such as exact MIP solvers. A similar strategy was suggested in earlier research on mesoscopic relocation modeling within free-floating car-sharing systems \citep{weikl20152}.




This paper focuses on the algorithm utilized in Stage 1. We explore a fast ranking-based method aimed at maximizing overall customer travel time. It makes decisions by assessing car availability in different locations, the expected demand probability distribution, and estimated trip lengths. The algorithm is equipped to make decisions under real-time constraints imposed by platform requirements. 

Evaluating the algorithm, we determine two benchmark boundaries within which the algorithm's outcomes should reside. The lower boundary is derived by executing simulations devoid of optimization. The upper boundary is defined by resolving the MIP problem formulation with the aid of an exact algorithm.

\subsection{Contributions}

{

To summarize our contributions are the following:
\begin{itemize}
    \item We propose a unified framework for one-way free-floating car sharing system supporting the assumed business model. It is based on three components: (1) a zoning algorithm, which determines geographically coherent zones  grouping locations with similar temporal supply and demand characteristics, (2) dynamic prediction of demand and the number of available vehicles, and (3) a fast greedy ranking-based algorithm for making decisions on vehicle relocation and scooter trips. 
    
    Although the approach consisting in applying zoning in FFCS has already appeared in  \citep{weikl20152}, our solution differs in the granularity of the zones, as well as in the methods of their definition and validation. Various solutions to the VeRP problem that combine demand prediction with relocation optimization have also been applied in station-based systems; however, the proposed solution pertains to FFCS and additionally employs a fast decision-making algorithm ready for deployment in an CS operator’s online service.
    
    \item We present a comprehensive pipeline that encompasses the division of the service area into zones, the validation of the zoning results, the tuning of algorithm hyperparameters within a simulation setting, and the evaluation and confirmation of algorithm outputs. This evaluation not only involves objective assessment but also considers factors like conflicts and the imbalance between the number of relocations and transits.
    
    \item     In addition to the previously noted benchmark boundaries, we present several performance benchmarks. These encompass an analysis of staff size, the influence of zoning, the utilization of diverse predictors, a comparison with a hybrid algorithm employing local optimization via a MIP model, as well as scalability considerations for varying numbers of zones. All experiments utilize real-world data sourced from Poland's leading car-service provider.
\end{itemize}



\subsection{Paper structure}

The structure of the article is organized as follows: Section~\ref{sec:state-of-the-art} offers a review of existing literature on fleet optimization for car-sharing systems. Section~\ref{sec:methodology} describes the overall methodology. It is followed by Section~\ref{sec:zones}, which discusses the development of an algorithm to divide the service region into zones, together with an analysis and validation of the zoning results. Following that, Section~\ref{sec:relocation-algorithm} provides the details of the relocation algorithm, including the procedure for its tuning and the associated results. Lastly, Section~\ref{sec:conclusions} presents a summary and proposes future research directions.

\section{Literature review}
\label{sec:state-of-the-art}
Car-sharing services (CSS) have emerged in response to urbanization and the growing demand for sustainable mobility solutions. Numerous studies have investigated different aspects of car-sharing. 
In \emph{descriptive} research, topics such as the impact on household vehicle ownership, contributions to urban sustainability, and characteristics of CSS consumers have been examined. 
Meanwhile, \emph{prescriptive} research focuses on resolving decision-support challenges such as developing business models, optimizing fleet management, predicting demand, and addressing vehicle imbalance issues. 

Readers are encouraged to consult surveys specifically pertinent to CSS ~\citep{ILLGEN2019193,GOLALIKHANI2021102280,WU2022100028} for an extensive overview of literature related to prescriptive analysis in CSS, along with the review \citep{Teusch2023} that discusses car-sharing within the broader context of urban mobility.

Two primary business models dominate the car-sharing industry \citep{MUNZEL2020243}: \emph{business-to-consumer} (B2C) and \emph{peer-to-peer} (P2P). In the B2C model, vehicles are privately owned and rented to consumers via a platform managed by a service provider. This model can be subdivided further based on operational mode: station-based, where vehicles are picked up and returned at specific stations,  or free-floating (FFCS), where vehicles can be picked up and dropped off anywhere within a designated service area. Another distinction within the model is between round trip (vehicles are returned to the pickup location) and one-way (vehicles can be returned to a different location). 

\subsection{Decision types in car-sharing operations}


Three levels of decision-making can be identified within the B2C framework.\citep{ILLGEN2019193}:

\begin{itemize}
  \item \emph{Strategic decisions}, such as the long-term planning of infrastructure and fleet sizing. This includes determining service areas and, for station-based systems, the optimal locations of pick-up and drop-off stations.
  \item \emph{Tactical decisions}, such as the allocation of fleet sizes and workforce across time and space, including staffing levels for vehicle relocation tasks, reservation strategies and pricing policies.
  \item \emph{Operational decisions} including setting the initial stock levels, maintenance planning and decisions  made in real time or near-real time, such as vehicle relocation and dynamic pricing.
\end{itemize}

Prescriptive analytics have been increasingly explored to support all these decision-making levels, using demand forecasts and contextual data to generate optimized actions. 
This paper specifically addresses operational decisions regarding vehicle relocation. 


\subsection{Predictive modeling for demand estimation}
\label{sec:demand-estimation}

Accurate demand forecasting is essential for ensuring vehicle availability and optimization of resource allocation.
Car-sharing systems have shifted from relying on predefined reservations to accommodating dynamic, stochastic demand.  Early studies used static table-based models or aggregated demand matrices. These were followed by statistical distributions calibrated to historical trip data. 
Kernel Density Estimation (KDE) emerged as a transparent, spatially aware method of estimating short-term demand. 
The study by \citep{ligaowang2022data} examined the efficiency of a vehicle relocation algorithm for station-based services across different demand models, employing fitted distributions such as Gaussian, Laplace, Poisson, and KDE. The optimization outcomes for the stochastic model revealed that KDE yielded the superior performance. 

Demand models focused on time series prediction employed statistical approaches such as ARIMA and SARIMA \citep{weikl2015,abbasi2022}, which are suitable for short-horizon forecasts. 

A large group of works incorporate supervised machine learning algorithms for demand prediction. Regression with Negative Binomial distribution was used in \citep{muller2017}. Poisson regression framework to detect demand shortfalls, aiding proactive fleet rebalancing, was proposed in \citep{burgin2023_spatiotemporal_model}. In \citep{alencar2021forecasting} various univariate and multivariate forecasting methods were tested, including ARIMA, SARIMA, Prophet, CatBoost, XGBoost, and LSTM. For short-term predictions, gradient boosting algorithms proved to be the most effective, while LSTM surpassed other models for long-term forecasting.


 \citet{daraio2020predicting} investigated short-term vehicle availability forecasting in FFCS using a variety of machine learning techniques. Their study revealed that non-linear models, such as SVM and tree-based regressors, significantly outperform linear baselines in highly dynamic environments. They also examined how prediction horizon and local urban features influence model accuracy, offering insights directly applicable to zone-based systems like ours.

Recent works incorporate supervised machine learning algorithms such as Random Forest, Gradient Boosting, and neural networks \citep{Giorgione2020,WANG2024681}, with several studies exploring spatial dependencies through GNNs and Transformer-based models \citep{LuoDuKlemmer2020,Shaheen2021}. 
The focus of the paper \cite{YU2020} is on evaluating the operational efficiency of electric, station-based car-sharing systems.
The study employs an LSTM model to predict short-term vehicle pick-up and drop-off patterns based on temporal characteristics such as the time of day, day of the week, and weather conditions.
Demand estimation was also performed through an agent-based simulation \citep{HEILIG2018151}.

Accurate demand forecasting is crucial for urban transportation. The review by \citep{Teusch2023} compiled literature on demand forecasts across different transportation modes such as car-sharing, ride-hailing, scooters, and bikes. It is important to highlight that predicting demand for fine-grained regions, considering both spatial and temporal distribution, presents significant challenges. Notable results were obtained in forecasting the aggregate demand for a vast area, such as the city of Vancouver, as demonstrated in \citep{alencar2021forecasting}. A similar problem was addressed in the context of demand prediction for a bike-sharing system \citep{MA2024129492},  where a prediction model combining CNN, LSTM, and the attention mechanism yielded an $R^2$ score of approximately 0.975. However, the prediction target encompassed the demand across the entire city of Washington. In contrast, \cite{LuoDuKlemmer2020} succeeded in predicting the fine-grained spatial demand distribution aggregated over long time periods to aid decision-making about the location of electric car stations.

A promising approach to accurate demand forecasting in fine-grained space-time grids involves incorporating inter-location flows into the model. This strategy, based on Graph Neural Networks, was adopted by \citet{FengLiuZhou20224}, yielding results that outperformed other methods.



\subsection{Vehicle imbalance}
\label{subsec:vehicle-imbalance}

Vehicle imbalance is a persistent operational challenge in one-way station-based and free-floating services. After finishing a trip, users can leave vehicles at a location that is different from where they started. In consequence, vehicles tend to accumulate in areas with low return probability, creating a spatial mismatch between supply and demand. 

Three main strategies commonly used to address vehicle imbalance can be identified \citep{GOLALIKHANI2021102280}. The first is \emph{staff-based relocation}, where personnel are tasked with transferring vehicles from stations or regions with an excess to those experiencing a shortage. The second strategy, known as \emph{user-based}, involves encouraging users to drop off vehicles in desired areas through mechanisms often through price-based incentives \citep{SaschaSAchulteVoss2014,STOKKINK2021230,LIU2022102884}. The third strategy focuses on \emph{trip selection}, involving decision whether to accept or decline trip demands based on their pricing and influence on the system's balance, all without resorting to relocation processes. Dynamic pricing falls into the last category. It can improve vehicle stock balance by offering variable prices depending on stock imbalance in the origin and destination areas. Although this solution is typical for ride-hailing services, it can also be applied to one-way car sharing.



Common methods for staff-based relocation typically relied on the use of mathematical programming, simulation, or a combination of both techniques. Mixed Integer Programming (MIP) and Mixed Integer Linear Programming (MILP) have traditionally been used to model optimal relocation under demand uncertainty \citep{ILLGEN2019193}. However, due to computational requirements, these approaches are often limited to small-scale or offline settings. Stochastic optimization methods and hybrid MIP-simulation frameworks have also been proposed to incorporate uncertainty, yet are rarely feasible in real-time operations. 

According to the survey conducted by \citep{GOLALIKHANI2021102280}, methodologies employing MIP or MILP are commonly applied to address the vehicle imbalance issue. It is important to highlight that when optimization is implemented across the entire horizon represented by the model, the algorithm also handles vehicle-user assignments, which aligns with the \emph{trip selection} strategy. As an alternative option, integrating MILP models with simulation, one can focus on short-term decisions regarding vehicle relocation \citep{santos2019finding,repoux2019dynamic}.

Other optimization algorithms used include Ant Colony Optimization \citep{HerbawiKnoll2016} Tabu Search \citep{CAI2022108005} and Adaptive Large Neighborhood Search (ALNS) \citep{LIU2023106220,EILERTSEN2024553}.  

Defining an optimization objective is necessary for solving the problem of vehicle imbalance through optimization. In works analyzed in \citep{GOLALIKHANI2021102280} goal functions included profit maximization, cost minimization, and maximization of the number of served requests. More recent articles, e.g. \citep{WANG2024681} base the objective on relocation distance and carbon emission.  

Integrating vehicle relocation with maintenance functions appears to be a straightforward expansion of the model. After conducting maintenance tasks, such as cleaning or recharging, it is logical to station a vehicle in an area of high demand. In \citep{TIAN2024121528}, the authors introduced such an integrated model that was subsequently reformulated as a capacitated vehicle routing problem. This was solved using a modified ant-colony algorithm.

In a different research direction explored in \citep{EILERTSEN2024553}, the integration of relocation strategies and dynamic pricing was examined. The authors introduced a two-stage stochastic model. In the first stage, decisions regarding relocations and pricing are made, while in the second stage, decisions concerning rentals are determined.


User-based relocation is another emerging field, in which users are offered financial or service-based incentives to drop vehicles in target zones. Typically, user-based relocation was analyzed with the use of simulation \citep{Kek2009,clemente2013vehicle,wagner2015data,STOKKINK2021230}. A recently proposed modular system supports agent-based simulation and various vehicle-sharing modes (see reference \citep{JimenezMerono02012025}). This software is designed to model and analyze a variety of problems, including station locations and various relocation strategies.


The article \citet{WEIDINGER20231380} provided empirical evaluations of vehicle relocation in FFCS for various booking strategies. The authors proposed a basic MIP model, which was then adapted to test various scenarios, such as time flexibility of requests, customer willingness to walk, fleet size, and vehicle relocation. The problems were solved with Gurobi and with a developed heuristic algorithm, which turned out to be on average three times faster. The findings presented reinforce the value of heuristic optimization under real-world constraints and validate the practical design of our ranking-based approach.

\subsection{Zoning and clustering}
\label{subsec:review-zoning-clustering}

The term 'zoning' refers to the division of the service areas into smaller disjoint zones. Although very often zones are not determined explicitly, the partition may be induced by the location of parking stations or charging stations for electric vehicles. The simplest imaginable partition is the Voronoi tessellation around their locations. 

Typically, decisions on station locations together with their capacity are based on observed or anticipated demand, as well as geographic features of the neighborhood. These features may include population, the presence of specific Points of Interest (POI), public transport alternatives, or prevailing land use. The importance of various factors influencing demand and station location was analyzed in \citep{ABBASI2021123846} using the Heckman selection model. Another recent study \cite{ZU2025104586} used Negative Binomial and Generalized Additive Mixed regression models to predict the usage of the car station based on geographical features and estimate their importance. 

In the study \citep{ChengChenDing2019} statistical models and machine learning algorithms were applied to assess demand across 58724 grid elements in Chengdu and propose station locations based on demand predictions and locations of existing stations. The station location problem was often modeled as multi-objective MIP \citep{BOYACI2015718,ChenZhangMaWang2019} or mixed non-linear integer programming \citep{HUANG20181} with an objective function to maximize satisfied demand. 
A different strategy was presented in \citep{XingChu2020}, where clustered demand data were utilized to optimize station placement using a whale optimization algorithm, aiming to minimize the overall walking distance.

Graph Neural Networks (GNN) combining geographical information and demand data were used in \citet{LIANG2023104241} to predict flows between bike sharing stations and based on these results recommend station locations. 

Zone-based approaches are an effective solution for reducing problem complexity in the case of FFCS. The concept of dividing service areas into distinct zones for discrete optimization at a macroscopic level was introduced in \citep{WEIKL2015206}. In this project, the Munich area was divided into 15 zones. The proposed vehicle relocation algorithm took inter-zone trips into account, as well as internal optimization of drop-off locations.

The details of a practical, decision support system that has been tailored for the Area Pricing Model can be found in \citep{Brendel2023}. The employed methodology involves clustering high-demand locations into larger zones and implementing a pricing strategy that incentivizes users to travel from low-demand to high-demand areas.

Integrating geographical data concerning zones with information on trip origins, destinations, and timings can be useful for relocation algorithms, especially when data is scarce. The study by \citep{FengKeYang2022} introduced an algorithm that merges a graph convolutional network with matrix factorization to forecast demand distribution from origins to destinations. While the research employed ride-hailing data, the findings are applicable to car-sharing services as well. Another important factor in relocation is the estimation of travel times. \citet{XuJonietzGupta2022} suggested an intriguing method to calculate the arrival time at a specific zone. Their approach involves constructing the shortest path in a graph of zones, where the weights are based on average travel times derived from actual data.

\subsection{User behavior and service design}


A comprehensive review of the literature in \citep{SHAMSESFANDABADI2022131981} highlighted that the integration of data on user behavior, intentions, and preferences is one of the four main areas of research in the context of car sharing.

The study \citep{RAMOS2023} examined the psychological factors that encourage or hinder the use of different car sharing business models. It identified the preferred models for specific trip purposes and highlighted the key motivations for using them. A Bayesian approach was used to analyze the survey data of car-sharing users in German cities, covering free-floating, round-trip station-based, and peer-to-peer models. 

An analysis of user preferences on vehicle rental fleets in Poland was conducted in \citep{turon2022rare}. A survey using the pairwise ranking method revealed that D-class electric cars were the most popular choice. User preferences can inform strategic choices in fleet modernization, such as dynamic pricing strategies and prioritization of vehicle relocation according to type.

The study \citep{Vejchodska2024} explored the differences in attitudes, motivations, and transport behaviors between car sharing users who have another car at home and those who do not. Individuals without access to a personal vehicle are more likely to use shared mobility options, although their overall mileage tends to be lower. 

\citet{MONTEIRO2023} conducted a study on the preferences of individuals for various car sharing models. The study analyzed the appeal of specific service features and incentives across different urban environments. A stated preference experiment was employed to collect data from Copenhagen, Munich, and Tel Aviv, with the objective of identifying the key factors that influence subscription choices. 
The findings indicate that, while certain elements, such as pricing and parking, are of universal importance, other preferences are strongly shaped by the specific local context.

Despite the great diversity of research on car sharing, in recent years the number of studies focusing on FFCS has been small. In particular, there are few works describing solutions consistent with our business model being shaped by the provider's assumptions, featuring key aspects like the lack of dedicated parking stations, a brief reservation period of maximum 15 minutes, and no predetermined trip destination. This was one of the factors that motivated us to conduct research in this area.

In recent studies, researchers’ attention has shifted to entirely different models and issues, such as combining early booking time and trip target specification \citet{WEIDINGER20231380}, user-based relocation \citet{STOKKINK2021230}, pairing relocation with electric vehicle charging \citet{TIAN2024121528} or even deployment of autonomous vehicles \citet{MA2017124}. These are important issues that may serve as a basis for building hybrid relocation models; however, an effective and online-capable solution for the basic staff-based relocation model in free-floating car sharing still remains an open research problem. As seen in numerous review articles, most methods focus on station-based services and are grounded in MIP models. Aspects of real-time operation and the potential for deployment within reactive services are analyzed rarely, and primarily in studies that extend beyond theoretical exploration to detail practical implementation efforts.


\section{Overall methodology}
\label{sec:methodology}

This paper proposes an integrated approach to solving the VReP in free-floating car-sharing systems. The method combines real-world data processing, spatial zoning, predictive modeling, and relocation decision to achieve  scalable optimization for dynamic fleet balancing. 

The overall framework consists of the following elements:

\begin{enumerate}
    \item \textbf{Service area zoning (Section~\ref{sec:zones})}: The urban area is partitioned into discrete zones to allow tractable modeling of vehicle flows and demand imbalances. This spatial discretization transforms a continuous urban space into a network of manageable relocation targets. Zoning is based on hexagonal tessellation  clustered into spatial units based on road distance and temporal similarity with respect to number of available cars and user demand. This step is crucial to aggregate the sparse demand signals and make the relocation problem solvable in discrete space.
    It should be noted, that the objectives of the devised zoning method are different of these discussed Section~\ref{subsec:review-zoning-clustering}. We expect that the resulting set of zones would harmonize with the needs of the relocation algorithm, whereas approaches discussed in the literature were more focused on optimization in the scope of strategic decision, like location of parking or charging stations.

    \item \textbf{Zoning validation (Section~\ref{subsec:zoning-validation})}: The obtained division into zones is then validated. The goal is to ensure a successful prediction of demand and supply within the zones. Validation steps include: (1) collecting real-world data on cars available for rent and events of user interaction (2) transforming them into time series (3) training machine learning algorithms (4) comparing their results averaged over zones.  

    \item \textbf{Predictive modeling}: Historical data is used to estimate vehicle availability and demand per zone for each time interval. The outputs are structured as multidimensional matrices (e.g., Poisson distribution parameters used to model demand $\Lambda[i,j,t]$, travel time $T[i,j,t]$). They further serve as input to the simulation and relocation algorithm. Temporal and spatial resolution is harmonized with the zoning structure.

    \item \textbf{Relocation algorithm (Section~\ref{sec:relocation-algorithm})}: Given predicted supply and demand across zones, the algorithm makes decision on relocating vehicles using staff and dispatching staff between zones. However, for evaluation purposes the algorithm is coupled with a simulation procedure that assigns vehicles to users at discrete time steps within the simulation horizon. The core of the method is a ranking-based optimization algorithm that balances demand coverage with relocation costs. The ranking uses zone-level inputs, which are only made possible by the prior zoning.

   \item \textbf{Evaluation (Section~\ref{sec:results})}: The decision-making process is evaluated using simulations over 96 time steps, corresponding to 15-minute intervals throughout a full operational day, using both historical and synthetic data. Multiple performance metrics (e.g. served demand, the balance between relocations and staff movements, computation time and scalability) are tracked and analyzed. 

 \end{enumerate}



\section{Division of the service area into zones}
\label{sec:zones}

The process of dividing the service area into zones consisted of three stages:
preparation and selection of data, application of an agglomerative clustering algorithm, and
validation of results.


\subsection{Preparation and selection of data}

We used data on the shape of the Kraków service area. Its boundary is shown in Fig.~\ref{fig:service-area-grid} marked with a blue line. The service area was covered with a hexagonal grid with a side length of 250 meters. It can be seen that the service area comprises a few detached zones located in the satellite cities Skawina and Wieliczka, as well as shopping centers, an office park and an airport. We decided to exclude them, as they are accessible by highways or roads with heavy traffic, which are not suitable for scooters. 

In the next step, we run the DBSCAN algorithm \citep{Ester1996} supplying hexagons centers as data points, with the parameter $\epsilon$ set to 500 m and the minimum number of neighbors equal to 3. The color-coded result of the clustering is presented in Fig.~\ref{fig:service-area-grid}. The hexagons belonging to the largest cluster marked in violet were selected for further processing. 
 
\begin{figure}[H]
    \centering
    \includegraphics[width=0.5\linewidth]{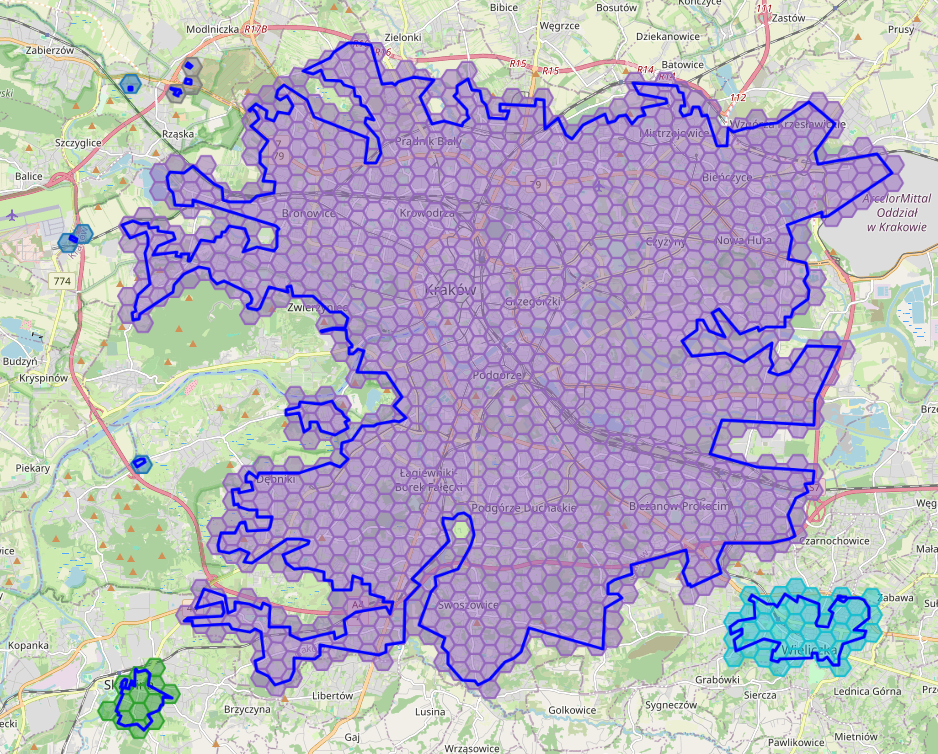}
    \caption{Service area covered by a hexagonal grid}
    \label{fig:service-area-grid}
\end{figure}

\subsection{Zoning algorithm}
\label{sec:zoning-algorithm}

A zoning algorithm is a method that is used to divide a geographical area into distinct zones based on specific criteria. These algorithms are commonly used in urban planning, transportation, telecommunications, and logistics to optimize resource allocation, service coverage, or spatial organization \citep{urbansci5030068}.

The goal of the presented zoning algorithm was to define areas that are suitable to optimize service operations through vehicle relocation. The determined zones should be characterized by similar geographical and temporal properties within their area. Moreover, several additional assumptions were made related to minimal or maximal zone size, as well as specific spatial relations, e.g. a zone should not be surrounded by another. 

We used an agglomerative clustering algorithm (cf. Algorithm~\ref{alg:agg_clustering}), which builds a hierarchy of zones by merging regions. It starts from an initial set of clusters, each comprising exactly one initial hexagon. In subsequent steps, the algorithm determines two candidate clusters to be merged based on their distance. The algorithm limits the size of the cluster, i.e. does not end at a single cluster, as well as handles special cases, in which a cluster is completely surrounded by another cluster. In such a case, the internal cluster is absorbed.  The algorithm was created as part of a master's thesis \citep{KamilRoman2023MscTh}.

\begin{algorithm}[H]
	\caption{Agglomerative clustering} 
	\label{alg:agg_clustering}
	\begin{algorithmic}[1]
		\Require
		\Statex $H=\{h_1,h_2,\dots h_m\}$ -- set of hexagons
		\Statex $d(A,B)$ - where $A,B \in 2^H$ -- distance function between sets $A$ and $B$
		\Statex $M$ - maximum cluster size
        \Statex
        \Statex \textit{Create $m=|H|$ initial clusters: }        
        \State $G \gets \{C_1,C_2,\dots C_m\}$, where $C_i=\{h_i\}$ 
        \While{ \textbf{$true$}}
            \Statex \hspace{0.5cm} \textit{Determine pairs of clusters that can be merged:}
            \State $P\gets\{ (A,B) \colon A, B \in G \wedge A\neq B \wedge |A| + |B| \le M\}$ 
            \If {$P=\emptyset$} 
                \State \Return $G$ 
            \EndIf
            \Statex  \hspace{0.5cm} \textit{Compute closest clusters $C_s$ and $C_r$:}
            \State $(C_r,C_s) \gets \arg \min_{(A,B)\in P} d(A,B)$
            \Statex \hspace{0.5cm} \textit{Join clusters $C_s$ and $C_r$:}
            \State $G \gets G \setminus \{C_s, C_r\} \cup C_{sr}$, where $C_{sr} = C_s\cup C_r$
            \Statex \hspace{0.5cm} \textit{Handle special case of clusters within other clusters:} 
             \While{ $\exists{C_i,C_j\in G \colon contains(C_i,C_j)}$}
             \State $G \gets G \setminus \{C_i, C_j\} \cup \{C_i \cup C_j\}$
             \EndWhile
        \EndWhile
	\end{algorithmic}
\end{algorithm}

The distance function used in the clustering algorithm is a weighted sum of five components (\ref{eq:distance-function}:
\begin{enumerate}
    \item Road distance $d_{rd}(A,B)$
    \item Distance based on comparison of road densities $d_{dns}(A,B)$
    \item Shape distance $d_{sh}(A,B)$
    \item Distance between temporal patterns of car presence $d_{cars}(A,B)$
    \item Distance between temporal patterns of user activities $d_{act}(A,B)$
\end{enumerate}

\begin{equation}
\label{eq:distance-function}
\begin{aligned}
d(A,B) &= w_{rd} \cdot \tilde{d}_{rd}(A,B) + w_{dns}\cdot \tilde{d}_{dns}(A,B) + w_{sh}\cdot \tilde{d}_{sh}(A,B) \\
&\quad + w_{cars}\cdot \tilde{d}_{cars}(A,B) + w_{act}\cdot \tilde{d}_{act}(A,B)
\end{aligned}
\end{equation}

\paragraph{Road distance} The road distance is an important clustering factor, as we assumed that a single cluster should not include areas separated by natural obstacles, such as rivers, or infrastructure elements, such as railway corridors, which usually require long detours. The distance is, in fact, a \emph{single linkage} distance computed according to formula (\ref{eq:dist-road}). The term $d_h(h_a,h_b)$ denotes the distance between the centers of hexagons (actually, projections of the centers on edges of the road network).    

\begin{equation}
\label{eq:dist-road}
    d_{rd}(A,B) = \min_{\substack{h_a\in A\\h_b\in B}}d_h(h_a,h_b)
\end{equation}

To speed up computations, the distance between non-adjacent hexagons is set to infinity. As there are 1173 hexagons computing distances between all of them would require about 680,000 shortest-path computations. If the number is limited to adjacent hexagons, considering that each has at most six neighbors, this number drops to about 3,500. 

The function $\tilde{d}_{rd}(A,B)$ appearing in formula (\ref{eq:distance-function}) denotes a distance normalized to the interval $[0,1]$:
\begin{equation*}
    \tilde{d}_{rd}(A,B) = \frac{d_{rd}(A,B)}{\max_{(P,Q)\in G \times G}\{d_{rd}(p,Q)\}}
\end{equation*}

\noindent Analogous normalization was applied to all distance components described below.

\paragraph{Distance between road densities} Information on the total length of the roads within the hexagons was collected from the Open Street Map (OSM) \cite{OpenStreetMap2024, Haklay2008}.  As hexagons are equal in size, it can easily be computed for all clusters. The density-based distance $d_{dns}(A,B)$ is calculated as the absolute value of the difference of densities for $A$ and $B$.

\paragraph{Shape distance} The shape distance was introduced to favor the spherical shapes of the clusters. It was frequently noted that employing a single linkage distance tends to yield clusters with chain-like forms. This led to the introduction of this distance measure to mitigate that effect. We used the Ward linkage method \cite{Ward1963}:
\begin{equation*}
 d(A,B)=\frac{|A|\cdot|B|}{|A|+|B|} \lVert \mu_A - \mu_B \rVert^2,   
\end{equation*}
where $\lVert \mu_A - \mu_B \rVert^2$ is a squared Euclidean distance between cluster centroids. 

\paragraph{Distance between temporal patterns of car presence and user activities} The method of calculation of $d_{cars}(A,B)$ and $d_{act}(A,B)$ is based on the Dynamic Time Wrapping (DTW) distance \citep{berndt1994, Salvador2007}.  
Each cluster $C_i$ is represented by a time series $X_i = (x_{it})$, the values of which are obtained by collecting information on the number of vehicles available in subsequent 15-minute intervals within a day and then aggregated over days. The distance between zones $A$ and $B$ is calculated as $dtw(X_A,X_B)$. For the distance $d_{act}$ related to the temporal pattern of user interactions, the difference lies only in the method of data collection. In this case, the number of events is collected when users interact with the system using the mobile application. 

Fig~\ref{fig:dtw-examples} shows examples of time series that represent the temporal patterns of car visibility and user interactions for two pairs of final zones. For better visibility, the second (green) plots were shifted down by 2 units.

\begin{figure}
    \centering
    \begin{tabular}{ccc}
         \includegraphics[width=0.45\linewidth]{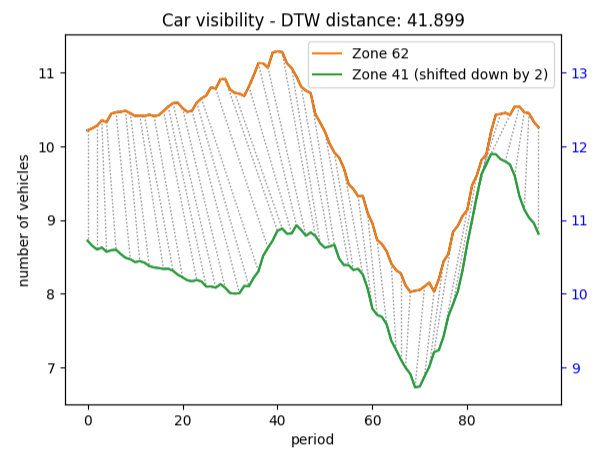}&&\includegraphics[width=0.45\linewidth]{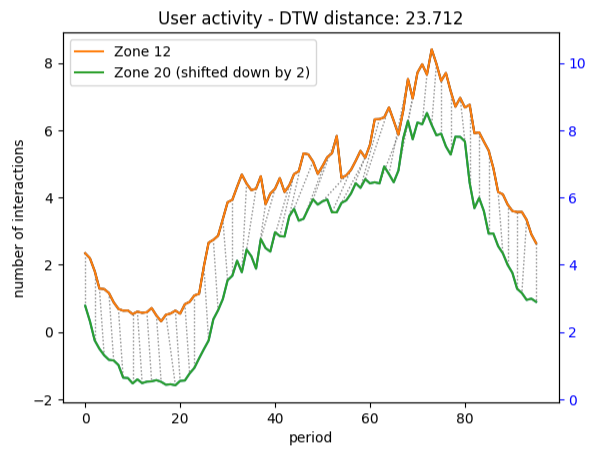}  \\
         (a)&&(b) 
    \end{tabular}
    
    \caption{Time series representing temporal patterns in selected zones related to (a) number of available cars (b) number of user interactions. Plots are shifted for better visibility. }
    \label{fig:dtw-examples}
\end{figure}

As we mentioned earlier, all distance functions are normalized to the interval $[0,1]$. Such a transformation was necessary because their values differed substantially. For example, the road distance between hexagonal centers was in the range 500m - 1000m, whereas the DTW distance, as shown in Fig.~\ref{fig:dtw-examples} are dozen times smaller. Choosing a value of 1 for all weights ensured an identical influence of the components on the aggregated distance value.

\subsection{Validation of results}
\label{subsec:zoning-validation}

The purpose of introducing zone division was to later use it in an optimization algorithm that improves vehicle availability through their relocation. The key factor in choosing the division was assumed to be good predictive properties, both in terms of the number of vehicles available in a zone and the estimation of demand.

Following the identification of the zone shapes, information was gathered for each zone concerning parked vehicles and user engagement with the system via the mobile app. A 15-minute time interval was selected for data discretization. Thus, for $i = 1, \dots, N$, $N$ time series $V_i = (v_{it})$ were created, representing the number of vehicles available in a given interval $t$, as well as $N$ time series $A_i = (a_{it})$ corresponding to user activity.

\subsubsection{Prediction of available vehicles}

We applied a typical approach for time series prediction, where the task is to predict the value $x_{t+h}$ of the time series $(x_1,x_2,\dots)$ given the previous $w$ observations $x_{i-w+1},\dots,x_i$. The parameter $h$ is the
prediction horizon and $w$ is the size of the sliding window.

Early trials indicated that predicting car visibility is fairly effective; nonetheless, accurately forecasting user interaction levels presents challenges. As a result, it was determined that the criterion for selecting the optimal zone division would be based on the reliability of predicting the number of available cars.

The following procedure was applied: for each $i=1,\dots,N$ zone, the time series $V_i$ were divided into a training set and a test set in the 70\% to 30\% ratio. Lasso algorithm was used as a regressor \citep{tibshirani1996regression}, the window parameter $w$ was set to 672, which corresponds to the use of historical values from one week. The experiments were carried out for two horizon values $h=3$ (90 min) and $h=6$ (90 min), and the mean value of the determination coefficient $\overline{r^2}$ was calculated for all zones and horizons. 

The tests were performed for 12 configurations of the distance function defined in the formula (\ref{eq:distance-function}) using weight values from the set $\{0.5,1,2\}$, which were intended to model the influence levels \emph{weak}, \emph{normal} and \emph{high}. The best configuration was determined by the $\overline{r^2}$ score. This optimal set was achieved with the following weight assignments: $w_{rd}=2, w_{dns}=1, w_{sh}=1, w_{cars}=1, w_{act}=1$, indicating a significant emphasis is placed on the road network distance.

The mean values of $r^2$ and other scores are summarized in Table~\ref{tab:zones-visibility-validation}. It is noticeable that for the 90-minute horizon, the results are somewhat less favorable (lower $r^2$ value and higher error scores), yet still demonstrate robust predictive performance.

\begin{table}[h]
    \label{tab:zones-visibility-validation}
    \centering
    \begin{tabular}{|l|ccc|ccc|}
        \hline
        & \multicolumn{3}{c|}{Horizon 45 min} & \multicolumn{3}{c|}{Horizon 90 min} \\
        \hline
        & mean & std & median & mean & std & median \\
        \hline
        r2&0.90 &0.13&0.92&0.82&0.13&0.84\\
        mse&0.38 &0.39&0.21&0.71&0.71&0.41\\
        rmse&0.55 &0.28&0.46&0.75&0.39&0.64\\
        maxe&4.79 &3.08&3.83&5.23&2.99&4.06\\
        med&0.21 &0.14&0.16&0.35&0.24&0.25\\
        mae&0.36 &0.21&0.29&0.52&0.30&0.43\\
                \hline
    \end{tabular}
    \caption{Summary of prediction measures calculated on test sets for 63 zones and two horizons: r2 -- determination coefficient, mse - mean squared error, rmse - root mean squared error, maxe - maximal error, med - median error, mae - mean absolute error }
\end{table}

Fig.~\ref{fig:r2_zones} presents maps depicting zones with color-coded values of determination coefficient $r^2$  for both horizons. Interestingly, predictions perform better in the city's outskirts. This can be understood by analyzing the data: vehicles parked in these areas often remain for several dozen hours, in contrast to the high vehicle turnover in the city center.

\begin{figure}[H]
    \centering
    \includegraphics[width=0.5\linewidth]{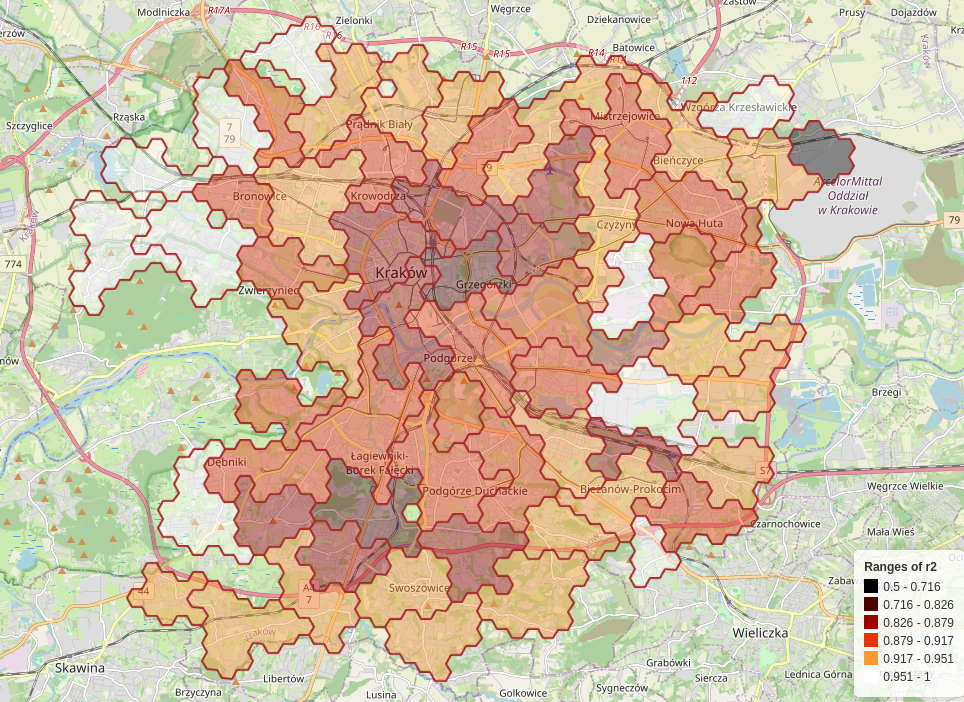}\\
    (a)\\
    \includegraphics[width=0.5\linewidth]{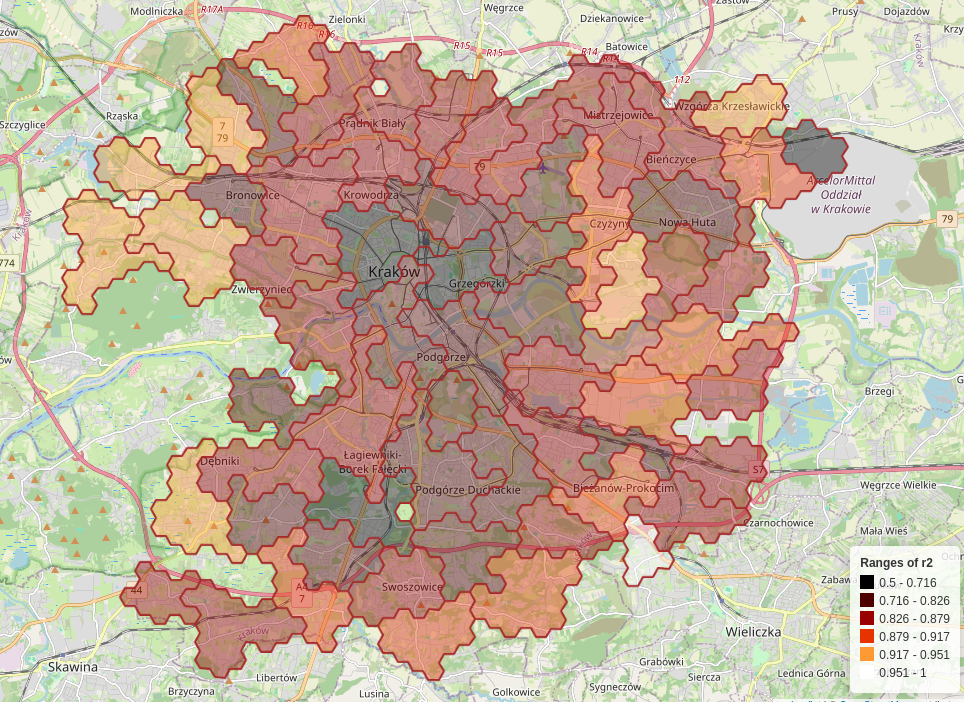}\\
    (b)
    \caption{Prediction of car presence. Ranges of $r^2$ scores obtained during testing (a) - horizon 45 min (b) - horizon 90 min}
    \label{fig:r2_zones}
\end{figure}

\subsubsection{Prediction of user interactions}
\label{subsec:user_interactions}

The number of user interactions through the mobile application indicates potential demand. The disparity between the demand and supply of vehicles to rent in a certain area is an important factor in making a decision about car relocation. 

Unfortunately, the distribution of user interaction is very uneven. Figure~\ref{fig:user-activities-total} displays data gathered over a year. A logarithmic scale enhances visibility. The numbers of interactions in one year range from 1 in the industrial area within former steel works to 140~000 at the Market Place, which is a popular tourist attraction. Typically, high numbers of user activities can be located at parking lots of shopping malls and at the airport (however, we decided to exclude the airport area from zoning, as it is located out of reach of scooters).

As mentioned earlier, we decided to exclude the prediction of the number of user activities from the evaluation of zoning results. 
The tests performed on time series $A_i = (a_{it})$ representing variations in the number of activities over time for particular zones did not yield satisfying results. Interactions within zones constitute rare events, the frequency of which changes significantly over time.

\begin{figure}[H]
    \centering
    \includegraphics[width=0.5\linewidth]{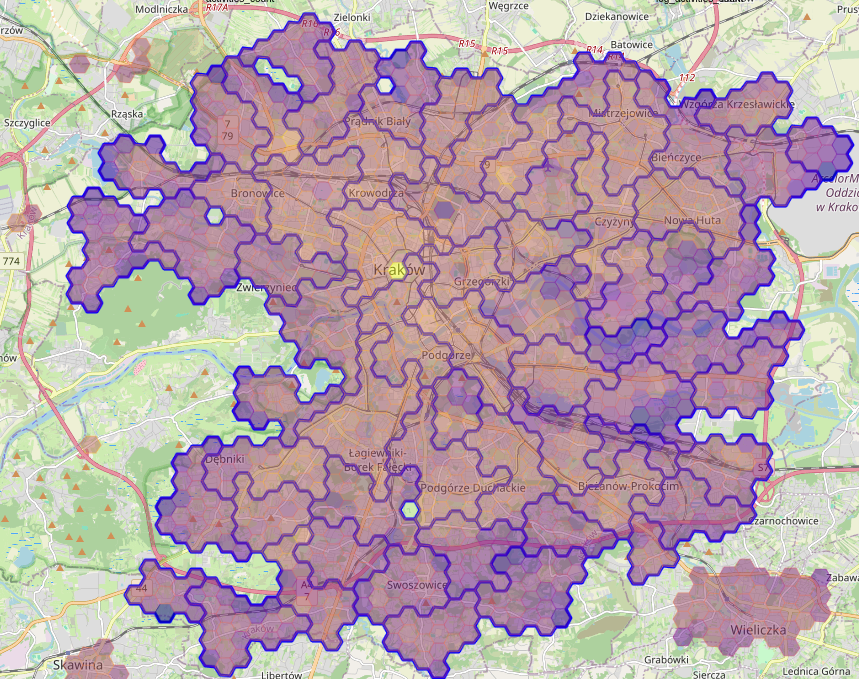}
    \caption{Distribution of numbers of user interactions during one year. Values range from 1 to 140,000. The logarithmic scale is used. }
    \label{fig:user-activities-total}
\end{figure}

Fig.~\ref{fig:activity_kde}a shows an example of the temporal distribution of 258 events collected during one day in a zone with an area of approximately $2.44 \text(km)^2$. As can be seen, events tend to concentrate at certain times of a day, which results in high variance of counts binned into 15 minute intervals. In fact, the dispersion index (variance to mean ratio) is very high VMR~$\approx 6$, which prevents effective prediction with various regression methods, as well as methods dedicated to the prediction of time series such as SARIMA, Negative Binomial, Zero Inflated Poisson \cite{Box2015, Hilbe2011, Lambert1992}. For all of them, the $r^2$ score was close to 0, and the predictions were practically equal to the mean. 

On the other hand, prediction for the entire service area can be made quite efficiently. For a one-hour horizon, typical scores for various algorithms were the following: $r^2\approx 0.95$, $MAE\approx 70$, $MED\approx 40$. This is due to the fact that the data are aggregated from about 60 times greater area, the number of active users is constant ($\approx 68 000$), and their actions are conditioned by the time of day, day of the week, weather, etc. This result is perfectly aligned with the observation on the efficiency of known demand prediction methods discussed in Section~\ref{sec:demand-estimation}.

Considering the sizes of the zones, the numbers of user interactions occurring locally cannot be efficiently predicted; however, the probability of such interactions exhibits significantly better properties. In Fig.~\ref{fig:activity_kde}a kernel density estimation (KDE) \cite{Silverman1986} of event probability is presented. A Gaussian kernel with a narrow bandwidth was used to emphasize the variation in time. 

Experiments carried out for all zones have shown that such probability density estimations can be very efficiently predicted, e.g., for one week window and two hour horizon application of simple linear regression yielded $r^2$ scores in the range 0.97 to 0.99. Fig.~\ref{fig:activity_kde}b shows typical results for time series representing one week. The plots of the true and predicted values in most places overlap.

One may be aware that predicting future values using KDE is debatable since the KDE method applied to the most recent observations inherently includes predictive mechanisms. However, the experiments conducted indicate that estimating the probability of vehicle demand can be a more reliable source of information than demand prediction itself, which typically returns values far from the ground truth.

\begin{figure}[H]
    \centering
    \begin{tabular}{ccc}
         \includegraphics[width=0.45\linewidth]{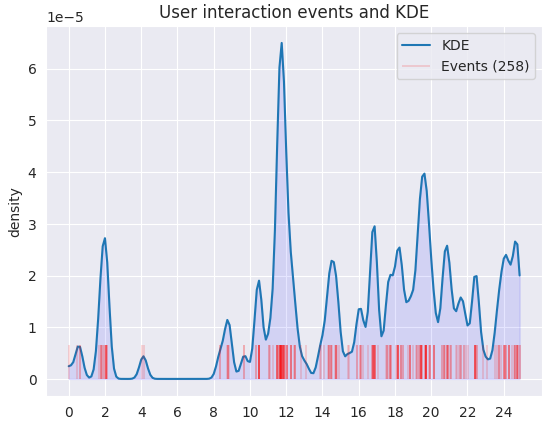}&&\includegraphics[width=0.45\linewidth]{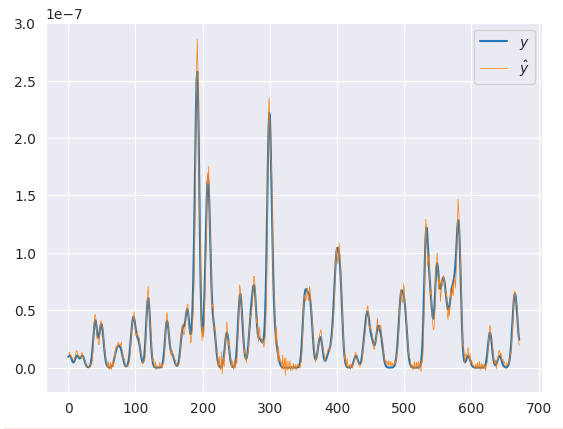}  \\
         (a)&&(b) 
    \end{tabular}
    
    \caption{Kernel density estimation applied to user activity: (a) events collected during one day and KDE aproximation (b) prediction of KDE during one week for 2 hours horizon}
    \label{fig:activity_kde}
\end{figure}

\subsubsection{Analysis of trips data}

The analysis of trips indicates that the zones differ significantly in terms of the number of trips started within them and their total duration. During the course of a year, trips with origins within the most popular zone amounted to a total of 8,196 hours. The average duration of travels per zone is 1,388 hours, but the median is 575, suggesting a high dispersion of values. Fig.~\ref{fig:zones-trips-time-flow}a shows the color-coded values of total travel times (the brighter, the higher). 

In turn, Fig.~\ref{fig:zones-trips-time-flow}b shows the flow of cars between zones during a year. For better visibility, only flows above 365 trips (one trip daily) are shown. Both maps suggest that there are actually about ten attractive zones with high service demand and total travel time, which can be increased by constantly relocating vehicles to these destinations. However, the relocation of vehicles from peripheral districts would reduce the availability of service in those areas, and the operating mode would become closer to a taxi service. Therefore, vehicle relocation should be carried out on a limited scale. 

\begin{figure}[H]
    \centering
    \begin{tabular}{ccc}
         \includegraphics[width=0.45\linewidth]{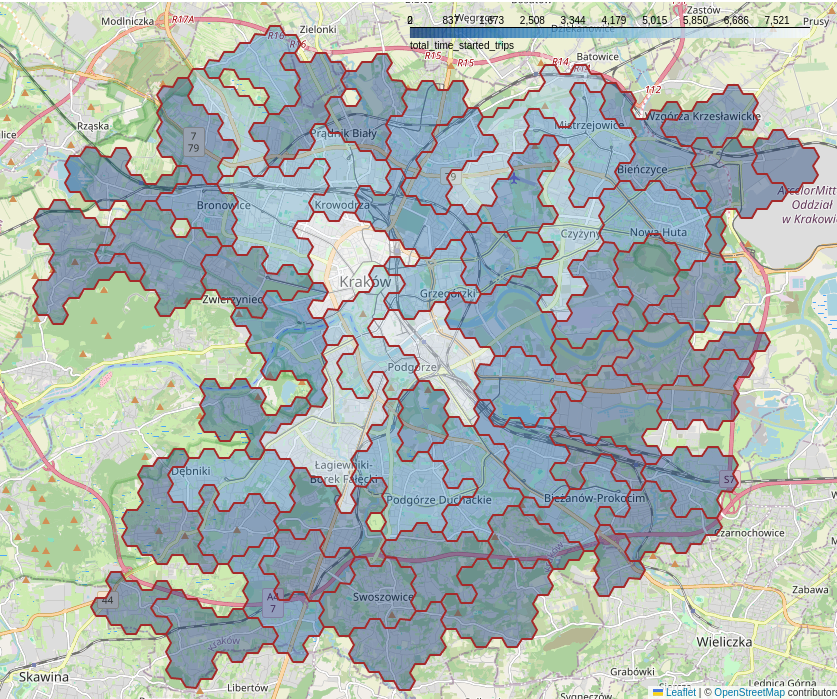}&&  
         \includegraphics[width=0.45\linewidth]{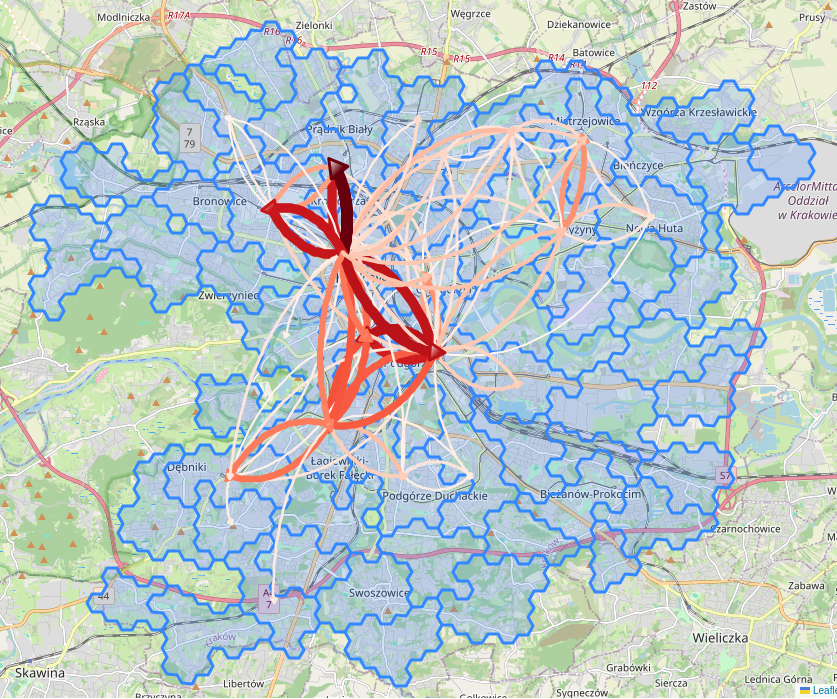}\\
         (a)&& (b)
    \end{tabular}
    
    \caption{Attractiveness of zones as vehicle relocation destinations (a) total time of trips starting in given zones (b) flow of vehicles between zones}
    \label{fig:zones-trips-time-flow}
\end{figure}

\section{Relocation algorithm}
\label{sec:relocation-algorithm}

Within this section, we examine a ranking-based algorithm designed to solve VReP, which is customized for the particular service model. 
Fig.~\ref{fig:simulation-relocation} presents the context diagram of the algorithm discussed. First, the algorithm is tightly coupled with its environment, which is either the simulation or the real world system. In both cases, the algorithm operates under the assumption that decisions allocating cars to clients are external and occur in the environment. The algorithm is responsible for two decision types: planning the relocation of vehicles by staff members or guiding them to a different location via scooter. 
Second, we opted to logically separate the forecasting components (\emph{vehicle availability predictor} and \emph{demand predictor}), which can operate based on both historical and recent data collected online within the real-world system or historical data optionally augmented by simulation.  A similar approach applies to \emph{travel time estimator}, which primarily relies on tabulated data but can be substituted by an online service.

It is important to highlight that the integration of the algorithm with the simulation is primarily for evaluation purposes. This embedding introduces certain constraints, such as time granularity, where decisions are made at intervals of 15 minutes. In a real-world scenario, relocation decisions are intended to be executed within a soft real-time deadline of a few seconds.

\begin{figure}[H]
{
    \centering
   \includegraphics[width=0.4\linewidth]{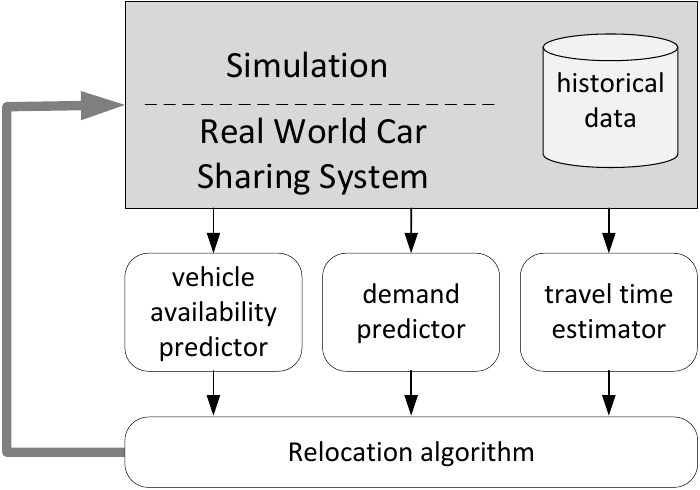}
    \caption{Simulation vs. real-world deployment}
    \label{fig:simulation-relocation}
}
\end{figure}

\subsection{Car sharing service model}

The analyzed model presupposes that trips occur among 63 zones (which also includes internal trips) without factoring in the specific spot where a vehicle is parked. Time is segmented into 15-minute intervals, implying that vehicle availability for clients, along with decisions concerning vehicle repositioning or crew movements on scooters, happen at set discrete moments.

Since high variability in travel times is observed throughout the day, a travel time matrix $T$ of size $N \times N \times P$ was constructed based on the data, where $N = 63$ (corresponding to the number of zones) and $P = 96$ (the number of discrete time points within a day). The element $T[i, j, t]$ is interpreted as the average travel time of a trip starting at time $t$ from zone $i$ to $j$.

The simulation of the model consists of executing a scenario. The scenario $s$ is defined as a tuple $s=(D,x_{v0},x_{s0})$, where $D$ is a demand matrix, $x_{v0}$ is a vector of initial vehicle locations, and $x_{s0}$ is a vector of initial staff locations. The demand matrix $D$ is of size $N \times N \times H$.
Its elements $D[i, j, t]$ are interpreted as the numbers of waiting trip demands to start at time $t$ from zone $i$ to $j$. The parameter $H$ denotes the simulation horizon. During experiments, we set $H=P$, although nothing prevents us from setting $H$ differently.

The model behavior is described with a state equation comprising state variables:
\begin{itemize}
    \item $x_v[i,t]$ - number of vehicles in zone $i$, ($i=1,\dots, N$) at time $t$, ($t=1,\dots,H$)
    \item $x_s[i,t]$ - number of staff members in zone $i$ at time $t$
\end{itemize}


and decision variables:

\begin{itemize}
    \item $u_v[i,j,t]$ - number of vehicles assigned to start a trip form $i$ to $j$ at time $t$
    \item $u_r[i,j,t]$ - number of vehicles assigned for relocation form $i$ to $j$ at time $t$
    \item $u_t[i,j,t]$ - number scooter transits form $i$ to $j$ at time $t$
\end{itemize}


The scenario execution is evaluated with a scoring function $Q(u^s,s)$. This is defined as the sum of the travel times within the simulation horizon $H$. The term $u^s$ stands for the scenario-dependent set of decision variables $u^s=(u_v,u_r,u_t)$: 

\begin{equation}
\label{eq:scoring-function}
    Q(u^s,s) = \sum_{i=1}^N\sum_{j=1}^N\sum_{t=1}^H u_v[i,j,t]\cdot T_{e}[i,j,t]
\end{equation}

\noindent $T_{e}[i,j,t]$ represents the travel time adjusted to the horizon length ($\Delta t$ is the length of the interval, i.e. 15 minutes) 
\begin{equation*}
T_{e}[i,j,t] =
\begin{cases} 
T[i,j,t], & \text{if } t+\lceil \frac{T[i,j,t]}{\Delta t}\rceil < H, \\
(H-t)\cdot \Delta t, & \text{in opposite case}.
\end{cases}
\end{equation*}

A scenario can be executed randomly or subject to an optimization algorithm that maximizes the scoring function $Q(u^s,s)$ directly, uses a surrogate objective function, or applies certain heuristics evaluated afterward with $Q(u^s,s)$.

To evaluate the model or optimization algorithm, we use a function $\mathcal{Q_S}$ defined according to equation (\ref{eq:scoring-multiple}) as the expected value of $Q(u^s,s)$ across all possible scenarios. This function is approximated by the sample mean following the approach known as the sample average approximation (SAA) \cite{Shapiro2009}, which builds a set of scenarios $S$, randomly sampling their parameters, and then executes them and computes the average score value. 

\begin{equation}
    \label{eq:scoring-multiple}
    \mathcal{Q_S} =\mathbb{E}(Q(u^s,s)) \approx \frac{\sum{s \in S}Q(u^s,s)}{|S|} 
\end{equation}

\subsubsection{Model calibration}
\label{subsec:model_calibration}

The demand matrix $D$ for a scenario is sampled from a Poisson distribution. According to Poisson distribution formula $p(k)$ - probability of $k$ events occurring in a certain interval is expressed by the formula:
\begin{equation*}
    p(k)=\frac{\lambda^k\cdot }{k!}e^{-\lambda}
\end{equation*}
The maximum likelihood estimate is of $\lambda$ parameter, which is the average number of events in the interval. As for each scenario, a demand matrix $D$ of size $N \times N \times H$ is sampled, all lambda constants are gathered in a matrix $\Lambda$ of the same size. Its elements are calculated on the basis of historical data summarized in two matrices:

\begin{itemize}
    \item $\overline{TR}[i,j,t]$ - average number of trips from zone $i$ to $j$ starting at time $t$
    \item $\overline{AC}[i,t]$ - average number of user interactions occurring in zone $i$ at time $t$
\end{itemize}

Elements of $\Lambda$ are calculated according to the formula (\ref{eq:big-lambda}). The first factor represents the ratio between the number of interactions and the trips started in a zone $i$. The second ensures that the average total number of demands is distributed among particular zones with a frequency based on registered trips.

\begin{equation}
    \label{eq:big-lambda}
    \Lambda[i,j,t] = \frac{\overline{AC}[i,t]}{\delta \cdot \sum_{j=1}^{N}\overline{TR}[i,j,T]}\cdot \overline{TR}[i,j,t]
\end{equation}

It should be mentioned that some elements of the $\overline{TR}$ matrix were equal to 0. To allow sampling demands corresponding to these elements even with a very low probability, a small smoothing factor $\alpha$ calculated as 
\begin{equation*}
\alpha = 0.1 \cdot \min\{TR[i,j,t]\colon TR[i,j,t]>0 \}    
\end{equation*}
 was added to all elements of the matrix.

The parameter $\delta$ in formula (\ref{eq:big-lambda}) is an additional scaling factor. In fact, a question arises about the ratio between the number of user activities and actual trips. It is clear that the number of interactions would always be greater or equal to the number of trips, as demand can only be partially met. However, not all interactions are made with the intention to rent a car, and sometimes users generate multiple events in a short time by checking if there are cars available nearby.  

In Fig.~\ref{fig:activity-vs-started-trips}a two plots are displayed: the first shows the number of rented cars during a day, the second number of user interactions divided by the scaling factor equal to 15. The relationships between values in subsequent time periods look realistic. At night, almost all demands end up with the assignment of a car, whereas during the day there is a gap, which can be explained by the lack of cars.

We conducted an experiment to choose the scaling factor $\delta$ consisting of computing the $r^2$ score and the DTW distance between time series representing the number of interactions scaled by $\delta$ and the number of trips started. As can be seen in Fig.~\ref{fig:activity-vs-started-trips}b, the greatest similarity was achieved for $\delta$ in the range 23-25. However,  verification revealed that these values were too high, as in some cases the demand would be smaller than the number of trips. Therefore, $\delta$ was set to 15, which ensured a safe distance over time. 

\begin{figure}
    \centering
    \begin{tabular}{ccc}
         \includegraphics[height=0.33\linewidth]{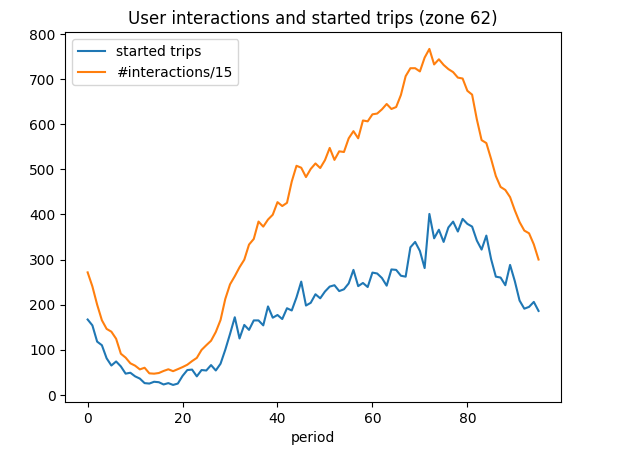}&&
         \includegraphics[height=0.33\linewidth]{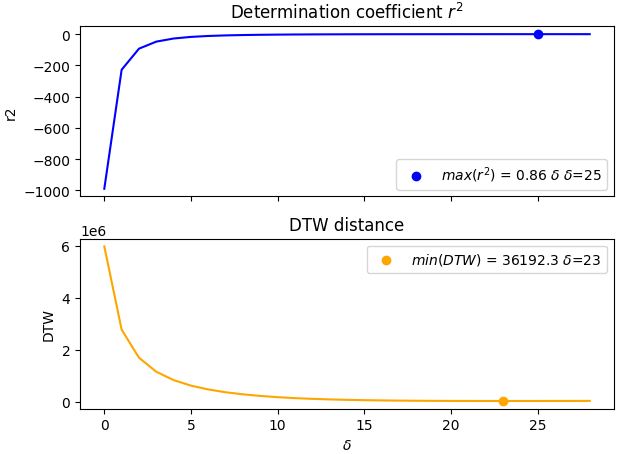}\\
         (a)&&(b)
    \end{tabular}
    \caption{Selection of scaling factor (a) interactions vs. started trips (b) plots of r\textsuperscript{2} and DTW for various scaling factors $\delta$}
    \label{fig:activity-vs-started-trips}
\end{figure}

\subsubsection{Service simulation}

The procedure \textsc{simulate\_assignment\_to\_clients} simulates the behavior of the car sharing system without optimization during one time step. Its flow is given in Algorithm~\ref{alg:simulate-asignment-to-clients}. The complete scenario is realized by repeating the procedure $H$ times. The calculation of the scoring function $Q(u^s,s)$ was omitted.

For a given zone $i$ the decisions $u_v[i,j,t]$ on the number of cars scheduled for trips to zones $j=1,\dots,N$ are made simultaneously by randomly sampling $k= \min\{x[i,t], \sum_j D[i,j,t]\}$ cars with probability:

\begin{equation*}
p(j)=\frac{D[i,j,t]}{\sum_{j=1}^N D[i,j,t]}    
\end{equation*}

\noindent This is performed within the \textsc{sample} function (line 7).

The next two steps (lines 8 and 9) specify the equations used to update the state variables. After scheduling cars to make trips, their number at a given location is reduced by the number of cars leaving. Based on decisions made, the number of cars arriving at chosen destinations is added to the state variables.

\begin{algorithm}[H]
	\caption{\textsc{simulate\_assignment\_to\_clients}($t,D,T,x_{v0}$)} 
	\label{alg:simulate-asignment-to-clients}
	\begin{algorithmic}[1]
		\Require
            \Statex $t$ - time moment
            \Statex $D$ - demand matrix
            \Statex $T$ - trip time matrix
            \Statex $x_{v0}$ - initial car locations
            \Statex 
            \If{t=1}
                \State $x_v[i,t]=x_{v0}$
            \Else
                \State $x_v[i,t]=x_v[i,t-1]$
            \EndIf
            \For{$t \gets 1$ to $N$} \Comment{$N$ is the number of zones}
                \State $u_v[i,\bullet,t]$ = \Call{sample}{$i,t,D,x[i,t]$}
                \State $x[i,t]\gets x[i,t] - \sum_j \sum_t u_v[i,j,t]$
                \State $\forall j: x[j,t+\lceil \frac{T[i,j,t]}{\Delta t}] \gets x[j,t+\lceil \frac{T[i,j,t]}{\Delta t}] + u_v[i,j,t] $
            \EndFor
	\end{algorithmic}
\end{algorithm}

\subsection{Mixed integer programming model}

We defined a full Mixed Integer Programming (MIP) model of car sharing service with car relocation. MIP approach is a popular choice for optimizing station-based car sharing services. There are many publications in which various MIP models have been defined with different levels of detail, often taking into account additional operations such as refueling, washing, and repairs \citep{FERRERO2018}. In the described applications, the problem was usually modeled as a stochastic MIP with two stage decisions, e.g. the first stage related to the number of cars and staff, and the second stage, where optimization of the car assignment is carried out. The objective function of Stochastic MIP is usually formulated as $f(x) = c^T\cdot x+\mathbb{E}(Q(x,s))$ and, as can be seen, compared to the scoring function defined in Equation (\ref{eq:scoring-function}), it includes an additional component $c^T\cdot x$, which usually expresses the cost of first-stage decisions (staffing, car leasing, etc.). Typically, the stochastic MIP approach was applied to problems with about 20 stations and a few hundred scenarios.

MIP model is defined as a set of equations linking decision and state variables (cf. lines 7 and 8 of Algorithm~\ref{alg:simulate-asignment-to-clients}) and inequalities expressing constraints, e.g. $\sum_j (u_v[i,j,t]+u_r[i,j,t] \le x[i,j]$ (the number of cars assigned to clients and scheduled for relocation is less than or equal to the number of cars present). 

MIP problems can be solved efficiently with optimization software tools such as Gurobi \cite{gurobi} or CPLEX \cite{cplex} that apply the branch-and-cut algorithm to give an exact solution. 

The primary purpose of developing the MIP model was not to use it as a direct optimization tool. Rather, we aimed to use it to evaluate the maximum scoring function values that could be achieved by employing algorithms that offer approximate solutions.

It should also be observed that in the MIP model the values of the variables expressing the assignment of cars to clients $u_v$ are not chosen randomly, but set by the solver. This could result in decisions constantly ignoring certain demanded destinations and preferring trips to zones with high flow of vehicles, cf. Fig.~\ref{fig:zones-trips-time-flow}b. Such decision rules fall into the category of \emph{trip selection} strategy om imbalance management (cf. Section\ref{subsec:vehicle-imbalance}). They can potentially be implemented in a real service,  e.g. by hiding vehicles from selected users based on their travel history, but would result in unequal access of customers to the service.

\subsection{Validation and comparison}
\label{sub:random-milp-validation-comparison}
The simulation model underwent validation using data from 2021 and 2022. This timeframe was chosen because, in that duration, the system operated with a minimal staff responsible solely for maintenance tasks. In the following years, passive relocation functions were implemented, alongside the introduction of some procedures for on-the-spot optimizations.

The tests confirmed the full consistency of the simulation model with real-world data. The total travel time during 24 hours averaged over 2000 scenarios was very close (with a gap of less than 2\%) to the daily mean. 

In the next step, we conducted an experiment aiming at comparing the scores obtained from the simulation of random vehicle assignment (without optimization) and MIP optimization. 

We generated data for 100 different scenarios through sampling demand and determining initial positions for vehicles and personnel. Each scenario was then run using the procedure \textsc{simulate\_assignment\_to\_clients} and optimized with the Gurobi solver, where the objective function matched the scoring function. The random assignment simulation yielded a mean trip time equal to 485.33 hours daily, while optimization of the MIP model with 10 staff members returned 580.44 hours daily. The difference was equal to 19.6\% and this number can be considered the upper limit on the improvements that can be made by various approximate algorithms  for those data (including the enhancements that can result from trip selection and preferential practices favoring particular clients).

For a single scenario, utilizing the Gurobi solver required around 60 to 70 seconds to finish on a laptop with 16GB of RAM and an i7 CPU. On the other hand, running a random scenario typically takes under 100 milliseconds. Most of the time spent running Gurobi was dedicated to the sequential process of defining the model, which requires substantial memory allocation. The parallelized nature of the optimization process itself ensures that it operates rather swiftly. The MIP solution was developed within engineering project \citep{DominikKikla2024EngTh}.

Fig.~\ref{fig:random-gurobi} illustrates the improvements obtained as a result of optimization (orange bars) with respect to the scores calculated for the random assignment procedure (blue bars).

\begin{figure}[H]
    \centering
    \includegraphics[width=0.7\linewidth]{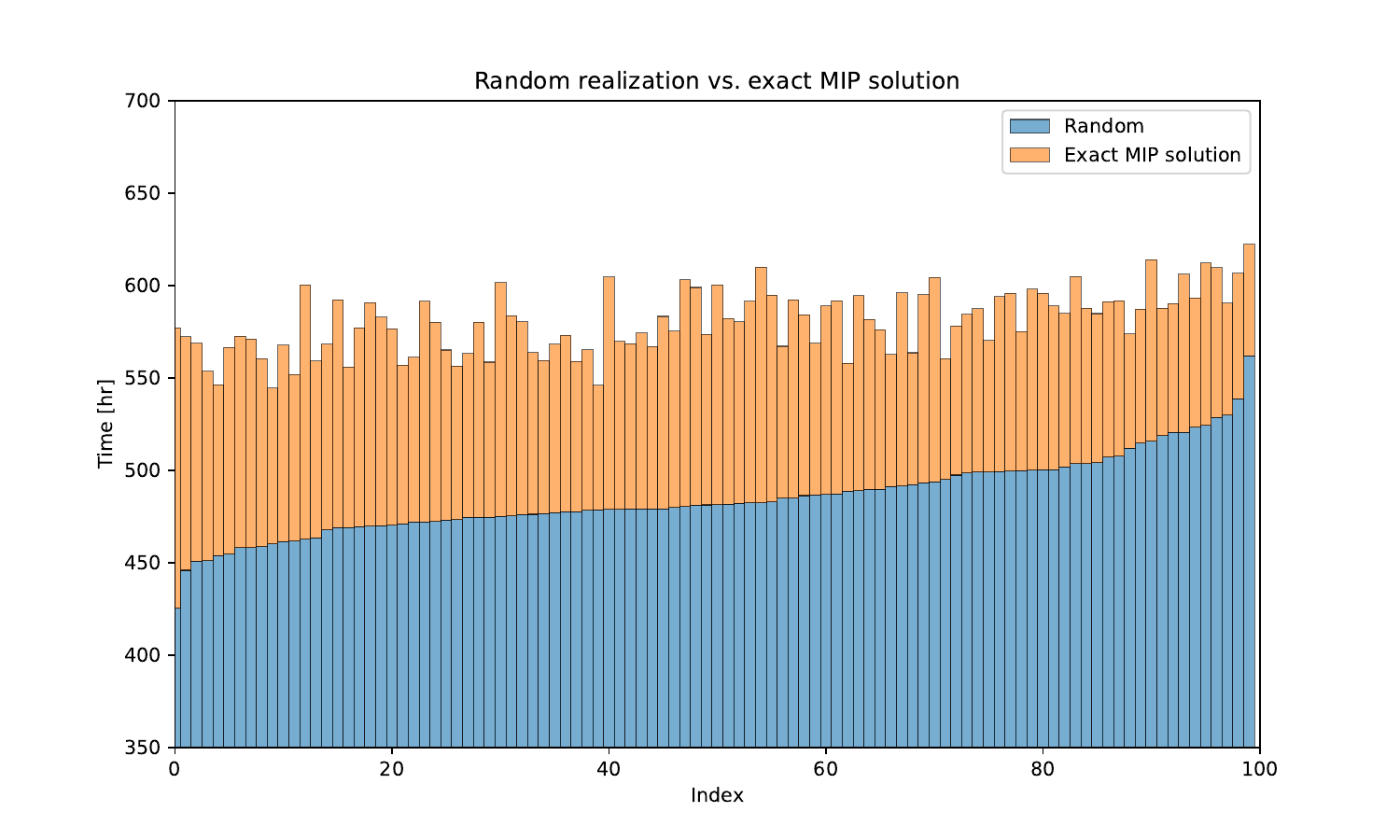}
    \caption{Comparison of total trip time obtained in a process random vehicle assignment and improvements made by exact algorithm solving MIP problem }
    \label{fig:random-gurobi}
\end{figure}

\subsection{Ranking based relocation algorithm}
\label{subsec:ranking-based-rel-algorithm}

This section outlines an optimization algorithm that employs greedy heuristics for scheduling car relocations and staff transfers using scooters. The overall approach is detailed in Algorithm~\ref{alg:high_level_optimization}. The execution of this algorithm involves two procedures, \textsc{schedule\_relocation} and \textsc{schedule\_transits}, which assess zones where staff are currently located and make decisions regarding vehicle relocations and scooter transits to other zones.
To improve readability, several procedure parameters were excluded, retaining only the algorithm's hyperparameters.

\begin{algorithm}[H]
	\caption{\textsc{VeRP\_simulation}} 
	\label{alg:high_level_optimization}
	\begin{algorithmic}[1]
		\Require
            \Statex $w_{tt}$ - trip time weight
            \Statex $w_d$ - demand weight
            \Statex $r_{th}$ - relocation threshold
            
            \Statex 
            \State Generate demand $D$ from $\Lambda$
            \State Generate initial vehicle location $x_{v0}$
            \State Generate initial staff location $x_{s0}$
            \For{$t \gets 1$ to $T$}
                \State \Call{simulate\_assignment\_to\_clients}{$t$} 
                \State \Call{schedule\_relocation}{$t,w_{tt},w_d, r_{th}$}
                \State \Call{schedule\_transits}{$t,w_{tt},w_d$} 
                
            \EndFor
	\end{algorithmic}
\end{algorithm}

Both procedures internally use concepts of \emph{vehicle availability predictor} and \emph{demand predictor}  (cf. Figure~\ref{fig:simulation-relocation}) that are assumed to return two predictions at time $t$  : 
\begin{itemize}
    \item $\hat{x}_v[i,t+h]$ - predicted number of cars present in zones $i=1,\dots,N$ at $t+h$
    \item $\hat{p}_d[i,t+h]$  - probability density of demand
\end{itemize}

\noindent Another common construct is to calculate the \emph{imbalance factor} $U$ for each zone as the weighted difference between supply and demand according to the equation (\ref{eq:imbalance}).
\begin{equation}
    \label{eq:imbalance}
    U[i] = \hat{x}_v[i,t+h] - w_d \cdot \hat{p}_d[i,t+h]\text{, for } i=1,\dots,N 
\end{equation}

\noindent The parameter $w_d$ is used to match the ranges of the number of cars in zones and the prediction of probability density (which is generally 2-3 orders of magnitude smaller).

Both method make decisions by calculating scores for zones ($R_r$ and $R_t$), organizing them into ranking lists, and systematically choosing items from the top as destinations. However, the calculation of these scores varies. 

For the \textsc{schedule\_relocation} procedure, $R_r$ scores are determined using formula (\ref{eq:ranking_rel}), and the zone with the smallest $R_r$ value is chosen for relocation. This approach prioritizes areas with lower (negative) $U$, signifying higher demand than vehicle supply, that can be accessed within short travel time $T[i,j,t]$. The weight $w_{tt}$ is employed as a scaling factor once again.

\begin{equation}
    \label{eq:ranking_rel}
    R_r[j] = U[j] + w_{tt} \cdot T[i,j,t] \text{ ,for } j=1,\dots,N
\end{equation}

Algorithm~\ref{alg:schedule-relocation} contains the pseudocode for the \textsc{schedule\_relocation} procedure. Parameter $w_{th}$ acts as a threshold. If the imbalance factor $U$ for a specific zone falls below its value, the vehicles should remain at place, since a low $U$ value signifies that demand exceeds supply.

\begin{algorithm}[H]
	\caption{\textsc{schedule\_relocation}} 
	\label{alg:schedule-relocation}
	\begin{algorithmic}[1]
		\Require
            \Statex $w_{tt}$ - trip time weight
            \Statex $w_d$ - demand weight
            \Statex $r_{th}$ - relocation threshold
            
            \Statex 
            \State Calculate $U$ according to (\ref{eq:imbalance})
            \For{$i \gets 1$ to $N$}
                \If {$U[i] < w_{th}$}
                   \State continue
                \EndIf
                \State Calculate $R_r$ according to (\ref{eq:ranking_rel})
                \State Calculate $M =\min \{x_v[i,t],x_s[i,t]\}$ 
                \For{$k \gets 1$ to $M$}
                    \State $r = \arg \min R_r$ \Comment{Select relocation target}
                    \State Set $u_r[i,t,t] \gets u_r[i,t,t] + 1$
                    \State Set $R_r[r] \gets R_r[r] - 1$
                \EndFor
            \EndFor
	\end{algorithmic}
\end{algorithm}

The operational principle used within \textsc{schedule\_tranists} is analogous. This approach involves calculating ranking scores $R_t$ using formula (\ref{eq:ranking_tr}), and the greedy assignment algorithm prioritizes locations with the highest scores, thus preferring areas with an excess of cars that are accessible with minimal travel time.

\begin{equation}
    \label{eq:ranking_tr}
    R_t[j] = U[j] - w_{tt} \cdot T[i,j,t] \text{ ,for } j=1,\dots,N
\end{equation}

Take note that the algorithm is not structured as an optimization process with a clearly specified objective function. Instead, it employs local greedy heuristics, and its outcomes are assessed post-execution using the scoring function delineated in (\ref{eq:scoring-function}).

The scores assigned to the destination zone in the \textsc{schedule\_relocati\-on} procedure are based on two factors: the predicted imbalance and the travel time. The last factor is particularly important as it helps eliminate time-consuming trips, such as during rush hour, between geographically close locations. Relying on travel times is further justified by the higher reliability of predictions when shorter travel times are promoted. We also believe that, in the context of Polish cities, this parameter plays a more significant role than distance in reducing CO\textsubscript{2} emissions and supporting sustainability.

It should also be mentioned that the scores used in the rankings can be easily extended with new features, such as incorporating priorities for the relocation of electric vehicles. This could be applied to the selection of the zone to which personnel on scooters are dispatched within the \textsc{schedule\_tranists} procedure. Currently, the algorithm does not differentiate between vehicle types as they are not individually represented within the simulation. The fleet is predominantly homogeneous, with electric vehicles comprising less than 5\% of the total number. As a result, accurately calibrating the simulation to reflect their usage would be challenging.

\subsection{Algorithm tuning}
The optimization algorithm has three hyperparameters that influence its behavior: 
\begin{itemize}
    \item $w_{tt}$ - trip time weight
    \item $w_d$ - demand weight
    \item $r_{th}$ - relocation threshold
\end{itemize}

During evaluations conducted using certain manually chosen values, two types of anomalies were identified. The first anomaly involved an unusually high frequency of relocations with only a few transits. Unexpectedly, this scenario was linked to a substantial rise in overall clients travel time, approximately by 14\%. Given the economic constraints, this setup was deemed impractical. In the second case, relocations were nearly absent, and the simulated staff activities primarily involved scooter transits. The resultant total travel times were comparable to those generated by a random simulation without optimization.

To choose hyperparameters that ensure both efficiency and realistic behavior, we used Optuna \citep{akiba2019optuna}, a hyperparameter optimization framework. To utilize Optuna, it's necessary to specify a goal function that accepts hyperparameters as inputs. During consecutive trials, Optuna constructs a probabilistic model of the goal function within the hyperparameter space and samples values that are more likely to yield favorable outcomes.

The applied goal function is specified in equation (\ref{eq:goal-optuna}).

\begin{equation}
    \label{eq:goal-optuna}
    f(w_{tt},w_d,r_{th}) = \overline{t} -10\cdot |R - T|
\end{equation}

\noindent where $\overline{t}$ is the total time of trips made by clients, $R$ is the total number of vehicle relocations, and $T$ is the total number of staff transits. Values of all variables are computed by executing the \textsc{Ranking\_based\_staff\_assignment} procedure (Algorithm~\ref{alg:high_level_optimization}).

The goal function is designed to maximize the total trip duration while maintaining a balance between the number of relocations and transits. This aligns with the anticipated mode of operation, where transits and relocations occur alternately: a staff member rides a scooter to a location with excess vehicles, relocates a vehicle to another site, and the cycle repeats.

The optimization was performed on 1000 samples (scenarios) for the following hyperparameter space: $w_{tt} \in [1e-4,1]$,  $w_d \in [20,1000]$ and $r_{th} \in [-30,30]$. The best hyperparameter values $w_{tt}=0.07$, $w_d=280.32$ and $r_{th}=-17.35$ were used in further experiments. 

The history plot showing the values of the objective function in subsequent trials is given in Fig~\ref{fig:optuna-history}.

\begin{figure}[H]
    \centering
    \includegraphics[width=0.7\linewidth]{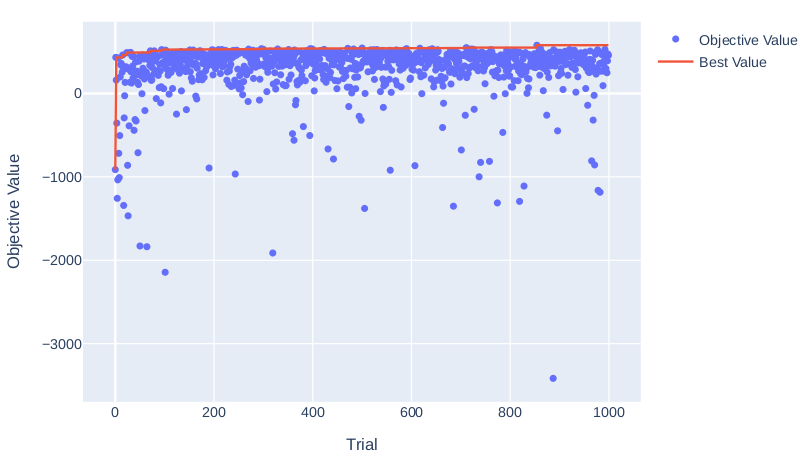}
    \caption{History of hyperparameters optimization with optuna}
    \label{fig:optuna-history}
\end{figure}

Observations from various trials show that values nearly as optimal as the best can be achieved, indicating that the algorithm is straightforward to fine-tune and yields consistent results across a wide range of hyperparameter settings.

\subsection{Optimization results}
\label{sec:results}

In an experiment, we ran 1000 scenarios spanning over 24 hours, with the number of cars set to 300 and a staff size of 7. Following the procedure outlined in Section~\ref{sub:random-milp-validation-comparison}, every scenario was executed using a random simulator before being submitted to an optimization algorithm. The average execution speed was about 4.2 scenarios analyzed in one second.

Throughout the trials, baseline predictors were employed, which utilized the current car counts along with cumulative demand from matrix $\Lambda$ to generate predictions:
\begin{equation*}
    \begin{aligned}
        &\hat{x}_v[i,t+h]=x_v[i,t] \\
        &\hat{p}_d[i,t+h] = \sum_{j=1}^N\Lambda[i,j,t+h]
    \end{aligned}
    \label{eq:two_lines}
\end{equation*}

Fig.~\ref{fig:hist-random-greedy} shows two histograms of the scoring function values obtained with a random assignment of vehicles without optimization (left) and using the described optimization procedure. The mean values and standard deviations are marked in the plots. The average total travel time through the scenarios was equal to 494.96 in the first case and 528.92 in the second. The algorithm increased the total trip time score by 6.86\%.

\begin{figure}[H]
    \centering
    \includegraphics[width=0.7\linewidth]{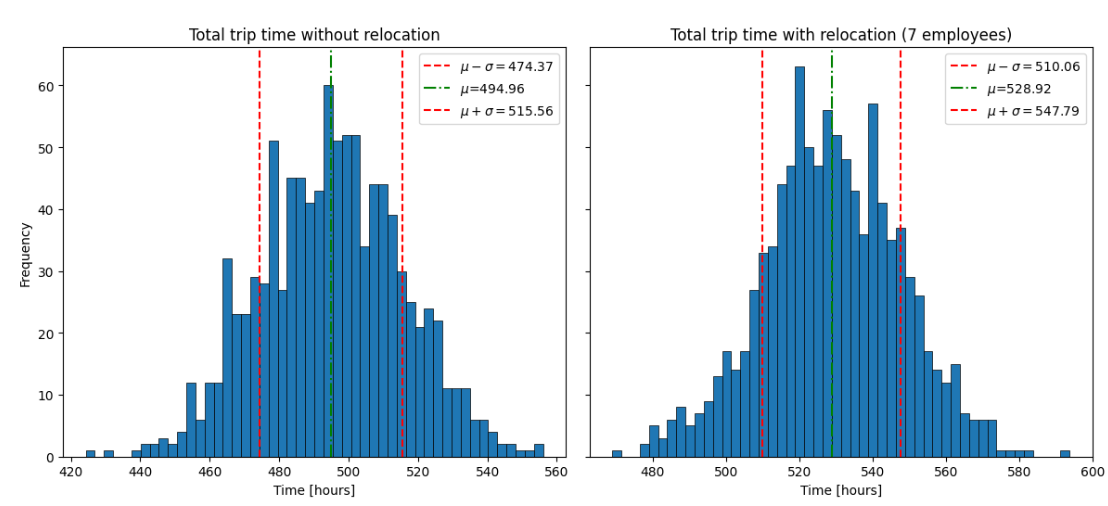}
    \caption{Histograms of trip times obtained for 1000 scenario executions (a) simulation of random behavior (b) combined simulation and optimization}
    \label{fig:hist-random-greedy}
\end{figure}

Table~\ref{tab:trips-analysis} gives details on the number and total time of trips, relocations and scooter rides. As can be seen, the average number of scooter transits and relocations were quite close, which indicates that the applied earlier tuning procedure was effective. Given that 7 employees consistently working throughout 24 hours equate to 21 Full Time Equivalent (FTE), we additionally computed the average durations and quantities of relocations and transits occurring within a single shift.

\begin{table}[H]
\centering
\begin{tabular}{lrrrrrr}
\toprule
 & $\overline{t}$ [hr] & Pct.$\overline{t}$ & $\overline{n}$ & Pct.$\overline{n}$    \\
\midrule
Trips & 528.92 & 86.69\% & 1022.78  & 85.89\%\\
Relocations & 81.18 & 13.31\% & 167.97 &  14.11\%\\
Relocations/FTE & 3.87 & N/A & 8.00 &  N/A\\
Total vehicles & 610.11 & 100\% & 1190.75 &  100\%\\
Scooter transits & 71.97 & 100\% & 152.19 &  100\%\\
Scooter transits/FTE & 3.43 & N/A & 7.25 &  N/A\\\bottomrule
\end{tabular}
\caption{Summary of trips, relocations and transits: $\overline{t}$  - average time over scenarios, Pct.$\overline{t}$ - percentage share of trips time wrt. all trip types   $\overline{n}$  number of trips, Pct.$\overline{n}$ - percentage share of trips wrt. all types}
\label{tab:trips-analysis}
\end{table}

\subsubsection{Conflicts}
By the term conflict, we describe the situation in which a staff member, after making a scooter transit to a certain location realizes that the vehicle which was intended to be relocated is missing. This behavior can be observed due to the assumption that client demands have greater priority than car relocation.

Information about conflicts is collected during the execution of the optimization procedure. During simulations, conflicts occurred with a frequency of 96\% with a mean value of 12.2. The histogram of conflicts (with omission of zero values) is shown in Fig.~\ref{fig:hist-conflicts}.  

\begin{figure}[H]
    \centering
    \includegraphics[width=0.5\linewidth]{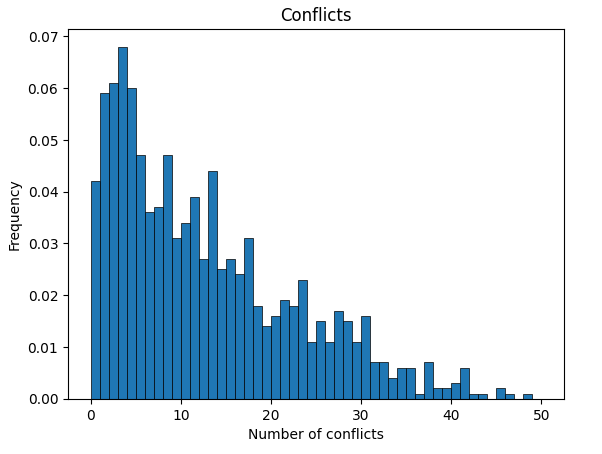}
    \caption{Histogram of conflicts (zero values were omitted)}
    \label{fig:hist-conflicts}
\end{figure}

\noindent Considering that staff size is equivalent to 21 FTE, on average 0.68 conflicts were encountered during the 8-hour shift. 

\subsubsection{Analysis of staff size}
The size of the staff is typically treated as an initial decision variable in Stochastic MIP problems, which is expressed as an extra cost in the formulation of the objective function. Since the main objective was presumed to be maximizing the total travel duration, this cost cannot be directly included in the scoring function formula. Nonetheless, we expect that analyzing how staff size influences scores could provide valuable business insights.

We conducted experiments for the staff size parameter in the range 0 to 20. Due to a lack of resources at staff size equal to 0 value, the behavior corresponds to random assignment of vehicles. Each time 1000 scenarios were analyzed. Results are summarized in Table~\ref{tab:num-staff-to-trips}, which apart from mean values gives also standard deviation and increase with respect to the 0 staff baseline.
The outcomes are presented visually as well in Fig.~\ref{fig:staff-trip-time}.

\begin{table}[H]
\centering
\begin{tabular}{lrrrrrr}
\toprule
 Staff size& $\overline{t}$ [hr] & $\sigma(t)$ [hr]& $\Delta t$ [\%] & $\overline{n}$ & $\sigma(n)$ & $\Delta n$ [\%]  \\
\midrule
0 & 494.89 & 20.31 & 0.00 & 956.04 & 37.26 & 0.00 \\
1 & 510.17 & 19.45 & 3.09 & 985.78 & 35.30 & 3.11 \\
2 & 515.50 & 19.05 & 4.17 & 996.88 & 34.37 & 4.27 \\
3 & 517.04 & 18.40 & 4.48 & 1000.11 & 33.30 & 4.61 \\
4 & 517.68 & 18.96 & 4.61 & 1001.90 & 34.66 & 4.80 \\
5 & 520.61 & 19.03 & 5.20 & 1007.64 & 34.04 & 5.40 \\
6 & 525.08 & 18.53 & 6.10 & 1015.61 & 34.04 & 6.23 \\
7 & 529.06 & 18.30 & 6.91 & 1023.63 & 33.35 & 7.07 \\
8 & 531.00 & 18.57 & 7.30 & 1027.44 & 33.44 & 7.47 \\
9 & 533.48 & 19.00 & 7.80 & 1032.71 & 34.82 & 8.02 \\
10 & 535.69 & 18.69 & 8.25 & 1036.70 & 33.56 & 8.44 \\
11 & 536.26 & 18.17 & 8.36 & 1037.96 & 32.89 & 8.57 \\
12 & 539.14 & 18.51 & 8.94 & 1043.96 & 32.78 & 9.20 \\
13 & 541.16 & 18.44 & 9.35 & 1046.43 & 33.81 & 9.45 \\
14 & 540.98 & 18.76 & 9.31 & 1046.71 & 33.82 & 9.48 \\
15 & 543.76 & 18.55 & 9.88 & 1051.87 & 33.84 & 10.02 \\
16 & 543.98 & 18.28 & 9.92 & 1052.73 & 33.59 & 10.11 \\
17 & 544.50 & 18.51 & 10.03 & 1053.52 & 33.66 & 10.20 \\
18 & 545.90 & 17.95 & 10.31 & 1056.98 & 32.35 & 10.56 \\
19 & 545.75 & 18.59 & 10.28 & 1057.02 & 33.78 & 10.56 \\
20 & 546.00 & 18.31 & 10.33 & 1056.30 & 32.96 & 10.49 \\
\bottomrule
\end{tabular}
\caption{The relationship between the number of staff members and the total trip time and the number of trips: $\overline{t}$ and $\sigma(t)$ - mean total trip time across scenarios and its standard deviation, $\Delta t$ - change with respect to random assignment case, $\overline{n}$ and $\sigma(n)$ - number of trips and standard deviation, $\Delta n$ - relative change}
\label{tab:num-staff-to-trips}
\end{table}

Predictably, as shown in Fig.~\ref{fig:staff-trip-time}, the curve eventually levels off at a 10\% increase. 
The MIP solution for a staff size of 10 resulted in an upper bound that was 19.6\% higher than the baseline. Under the same conditions, the investigated VReP  algorithm produced an 8.44\% increase, representing a notably smaller outcome. It is important to note, however, that the algorithm was integrated with a simulation modeling random assignment of cars to clients, whereas the MIP model, with prior demand knowledge, optimized these decisions as well.

The analysis of the collected data indicates that even with a limited staff size of 3-4 people (equivalent to 9-12 full-time employees), it is feasible to improve the scores by approximately 4\%.

\begin{figure}[H]
    \centering
    \includegraphics[width=0.7\linewidth]{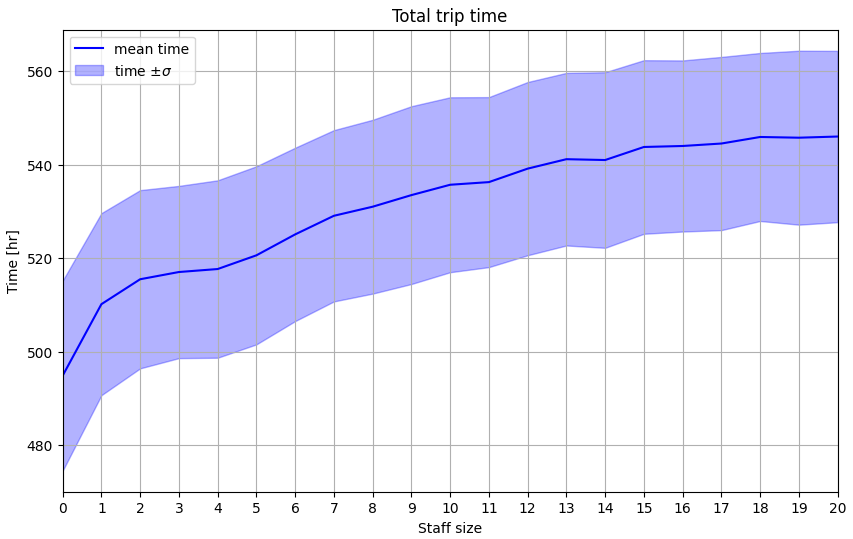}\\
    \caption{The dependency of total trip time on size of staff performing vehicle reloction. Standard deviation values computed across scenarios show in the plot indicate stable algorithm operation}
    \label{fig:staff-trip-time}
\end{figure}


\subsubsection{Impact of zoning}
\label{subsec:impact-zoning}

In Section~\ref{sec:zoning-algorithm}, we described the zoning algorithm, which internally incorporated a distance function composed of multiple components (see Eq.~\ref{eq:distance-function}). In particular, among these was the distance between temporal demand and supply patterns. This enabled the identification of areas with similar characteristics, which at specific times exhibited either an excess of rental vehicles or a scarcity. Ignoring these patterns might lead to the unintentional aggregation of locations with contrasting demand and supply trends, potentially canceling each other out.

To illustrate the behavior and evaluate the impact of the suggested zoning strategy on the performance of the relocation algorithm, we conducted an experiment,  which involved creating 63 zones using various clustering algorithms that considered only the Euclidean distance between the grouped locations and then executing the relocation algorithm.

During the tests, 200 scenarios with identical demand were generated, and the results of random execution without optimization were then compared with those obtained using the ranking-based vehicle relocation algorithm.

The outputs of the experiments are shown in Table~\ref{tab:impact-zoning}. As demonstrated, for the clusters determined by the zoning algorithm discussed in Section~\ref{sec:zones}, the most significant increase in both the number of trips and the duration of the trip can be observed after applying the ranking-based relocation algorithm. No matter the clustering technique used, the observable indicators remain stable, with the number of relocations nearly matching the number of transits, and the number of conflicts being comparatively low.

\begin{table}[H]
\centering
{\footnotesize
    
\begin{tabular}{lrrrrrrrrr}
\toprule
clustering & $\overline{n_R}$ & $\overline{n}$ &  $\Delta \overline{n}$ &  $\overline{t_R}$ [hr] & $\overline{t}$ [hr] & $\Delta \overline{t}$ & reloc. & trans. & conf. \\
 &  &  &  &  &  &  &  &  &  \\
\midrule
zoning & 960.65 & 1025.90 & 6.79\% & 497.21 & 530.38 & 6.67\% & 163.94 & 151.16 & 14.77 \\
kmeans & 972.99 & 1009.08 & 3.71\% & 487.61 & 505.31 & 3.63\% & 125.81 & 107.00 & 10.87 \\
bkmeans & 975.42 & 1010.50 & 3.60\% & 488.73 & 505.93 & 3.52\% & 121.88 & 104.28 & 23.03 \\
ac-avg. & 974.26 & 1007.51 & 3.41\% & 488.38 & 504.68 & 3.34\% & 116.65 & 111.94 & 20.52 \\
ac-compl. & 985.21 & 1018.41 & 3.37\% & 493.01 & 509.11 & 3.27\% & 110.78 & 113.81 & 29.11 \\
ac-ward & 984.07 & 1009.77 & 2.61\% & 493.41 & 506.03 & 2.56\% & 115.79 & 112.93 & 20.60 \\
\bottomrule
\end{tabular}    
}
\caption{Comparison of impact of clustering algorithm on the performance of vehicle relocation algorithm. $\overline{n_R}$ - number of random trips, $\overline{n}$ - number of trips after optimization, $\overline{t_R}$ - mean time for random trips, $\overline{t}$ - mean time after optimizations. Clustering methods: zoning - proposed in Section~\ref{sec:zones}, bkmeans - Biscecting k-means, ac-avg, ac-compl, ac-ward - agglomerative clustering with average, complete and Ward linkages. }
    \label{tab:impact-zoning}
\end{table}

As can be seen, the number of trips without optimization (column $\overline{n_R}$) is slightly different for each clustering method. This is a side effect of demand engineering - a small smoothing value is added to each element of the demand matrix (see Section~\ref{subsec:model_calibration}). Thus, the variation in the number of trips and their duration, as compared to random execution, provides more insight than the individual values themselves.
We performed one-way ANOVA tests on $\Delta n$ and $\Delta t$ values which confirmed differences within the group of clustering methods.  The post-hoc Tukey HSD pairwise analysis revealed distinctions between the zoning algorithm and each of the clustering techniques. No statistically significant differences were found among k-means, bisecting k-means, and agglomerative clustering using average and complete linkage. Nonetheless, the outcomes for agglomerative clustering with ward linkage were notably poorer, as indicated by a p-value less than 0.05.

\subsubsection{Benchmarking predictors}
\label{subsec:benchmarking-predictors}

In Section~\ref{subsec:ranking-based-rel-algorithm}, where we introduce the algorithm, we noted that it internally employs the idea of predictors to estimate the numbers of available vehicles and anticipated demand. This strategy supports a modular architecture, which can be adjusted during deployment to fulfill particular requirements. 

This section intends to evaluate how well the vehicle relocation algorithm performs with different predictors of vehicle availability. The count of vehicles within different zones acts as state variables derived from simulations, which subsequently inform predictions. Since demand events are not simulated, historical data is utilized to estimate demand.
It should be noted that during the experiments, the prediction concerns several dozen zones, and multiple simulation scenarios are carried out, which excludes the use of algorithms that require lengthy training or return predictions after a long delay. Algorithms capable of processing data streams would be preferred.

The following predictors were used during the experiments:
\begin{itemize}
    \item \emph{last observation} - a very simple predictor that returns the last observation
    \item \emph{ma4} and \emph{ma6} - moving average predictors with windows of 4 and 6 samples (60 and 90 minutes) respectively 
    \item \emph{lasso} and \emph{ridge} - these are linear predictors computing target variable according to the formula $y_{t+h}=\sum_{i=1}^n w_i\cdot x_{t-i-1}+w_0$ We used past $n=672$ values (equivalent of one week) to predict value in $h=2$ horizon (30 minutes). The algorithms differ in the type of regularization used during training.  
    \item \emph{xgboost} -- a tree-based ensemble regressor configured in the analogous way as lasso and ridge.
\end{itemize}

Time-series data were collected for each zone and separate lasso, ridge, and xgboost models were developed and validated. These models were saved for further use in simulation experiments. Given that the models employ a window of 672 samples, but the anticipated simulation horizon produces merely 96 data samples, predictions were made using merged real-world and simulation data. The initial three predictors (last observation, ma4, and ma6) did not need training; nonetheless, they were evaluated using the same datasets as the other predictors. Table~\ref{tab:predictor-benchmark} presents the average r\textsuperscript{2} scores. Unexpectedly, the simplest \emph{last observation} predictor is among those with the top scores, while the lowest scores were obtained by ma6 and xgboost.

All configurations were tested on 100 identical scenarios with staff size set to 7. Results are summarized in Table~\ref{tab:predictor-benchmark}. It can be observed that the algorithm results are quite similar and the algorithm setup (hiperparameters obtained during callibration) yield stable results (number of relocations is roughly equal to the number of tranists, number of conflicts is not high).

\begin{table}[H]
    \centering
{   
    \begin{tabular}{lrrrrrr}
    \toprule
     predictor & r2 & trips & trip time & relocations & transits & conflicts \\
    \midrule
    last observation & 0.94 & 1018.70 & 526.99 & 169.06 & 154.44 & 11.99 \\
    ma4 & 0.91 & 1012.09 & 523.56 & 158.72 & 158.77 & 13.11 \\
    ma6 & 0.89 & 1010.91 & 522.83 & 157.83 & 158.92 & 13.46 \\
    lasso & 0.92 & 1011.29 & 523.29 & 154.88 & 160.20 & 16.30 \\
    ridge & 0.94 & 1011.40 & 523.25 & 159.89 & 160.05 & 13.04 \\
    xgboost & 0.90 & 997.10 & 515.88 & 148.25 & 161.66 & 18.34 \\
    \bottomrule
    \end{tabular}
}
    \caption{Average values obtained during benchmarks (r2 averaged over zones other scores over scenarios)}
    \label{tab:predictor-benchmark}
\end{table}

A one-way ANOVA test was conducted, confirming the presence of statistically significant differences among the configurations using different predictors (F = 5.281, p=0.000093). Subsequently, a post-hoc pairwise Tukey HSD test was performed, which identified xgboost as differing from the other predictors (confirmed by p-values in range between 0.008 and 0.015). These findings are consistent with the data in Table~\ref{tab:predictor-benchmark}.

\subsubsection{Local optimization}
\label{subsec:local-optimization}

The algorithm described in Section~\ref{subsec:ranking-based-rel-algorithm} executes sequentially three ations in each time steps:
\begin{enumerate}
    \item Assigns vehicles to clients (cf. Algorithm~ \ref{alg:simulate-asignment-to-clients})
    \item Decides on vehicle relocations (procedure  \textsc{schedule\_relocation}, cf. Algorithm~\ref{alg:schedule-relocation})
    \item Unengaged staff members are directed on scooters to zones overstocked with free vehicles (procedure \textsc{schedule\_transits})
\end{enumerate}

In procedures \textsc{schedule\_relocation} and \textsc{schedule\_transits} first imbalance factors $U[i]$ are calculated for each zone $i$ and then ranking list $R_r$ and $R_t$ taking into account travel cost. Then a fast greedy procedure is applied that makes a decision: relocate a vehicle from zone $s$ (where the personnel member is present) to zone $t$ (with high demand). Similar decisions, but in opposite directions, are made on scooter moves from $s$ to $t$.

A question emerges regarding the extent of improvement achievable by substituting the greedy approach with an optimization procedure, preferably providing exact solution to Mixed Integer Programming (MIP) model.

To answer this we developed variants of procedures applying optimization locally, at each time step. 
The procedure \textsc{schedule\_relocation\_mip} makes relocation decisions $u_r$ by solving the optimization problem with a goal function  defined by equation (\ref{eq:goal_relocation_mip}. The function $f_r(U_r)$ considers the reduction of the imbalance $U$ along with the travel costs.

\begin{align}
    U'[k] &= U[k] - \sum_{j=1}^{N}u_r[k,j,t], &\text{for } k=1,\dots,N\\
    U''[k] &= U'[k] + \sum_{i=1}^{N}u_r[i,k,t], &\text{for } k=1,\dots,N\\
    C(u)&=\sum_{i=1}^n\sum_{j=1}^n u[i,j,t] \cdot T[i,j,t] \label{eq:travel_cost}\\
    f_r(u_r)&=\sum_{k=1}^nU''[k]\cdot \mathbf{1}(U''[k]<r_{th}) - w_{tt}\cdot C(u_r) 
    \label{eq:goal_relocation_mip}
\end{align}

A similar approach was applied, while developing the \textsc{schedule\_transits\_mip} procedure. The goal function $f_t(u_t$ prefers scooter transits to closely located zones with high $U$ values.

\begin{equation}
\label{eq:goal_transits}
    f_t(u_t) = \sum_{j=1}^N U[j]\cdot \sum_{i=1}^N u_r[i,j,t] - w_{tt}\cdot C(u_t)
\end{equation}

In our experiments, Gurobi was employed as the MIP solver. Each scenario, comprising 96 steps, required 192 local optimization tasks to be executed. Given the relatively small size of these local problems, Gurobi proved to be highly efficient. The average time taken to execute a single scenario was approximately 12.5 seconds.

The tests were carried out on 200 scenarios. For each scenario we collected outcomes of three algorithms: simulation without relocation, ranking based relocation  and local optimization using MIP model. Summary results are presented in Table~\ref{tab:rb_vs_local_mip}.

\begin{table}[H]
{ 
    \centering
\begin{tabular}{p{1cm}rrrrrrr}
\toprule
algorithm & $\overline{n}$ & $\Delta \overline{n}$  & $\overline{t}$ & $\Delta \overline{t}$  & reloc. & transits & conflicts \\
\midrule
NoOpt & 953.78 & NaN & 493.51 & NaN & 0.00 & 0.00 & 0.00 \\
RB & 1019.18 & 6.86\% & 526.98 & 6.35\% & 168.90 & 152.72 & 11.46 \\
MIP & 1037.09 & 8.73\% & 536.08 & 7.94\% & 301.06 & 119.12 & 0.01 \\
\bottomrule
\end{tabular}    \caption{Algorithm results averaged over 200 scenarios: $\overline{n}$ - average number of trips, $\overline{t}$ - average time in hours. Algorithms: NoOpt - no vehicle relocation, RB - ranking based, MIP - local optimization at each time step }
    \label{tab:rb_vs_local_mip}
}
\end{table}

Figure~\ref{fig:boxplot-ranking_based_mip} displays distribution of trips count and time for specific algorithms. The diferences between them were confirmed by one way ANOVA test and pairwise Tukey HSD tests, which yielded very small p-values for each pair.

\begin{figure}
    \centering
    \includegraphics[width=0.5\linewidth]{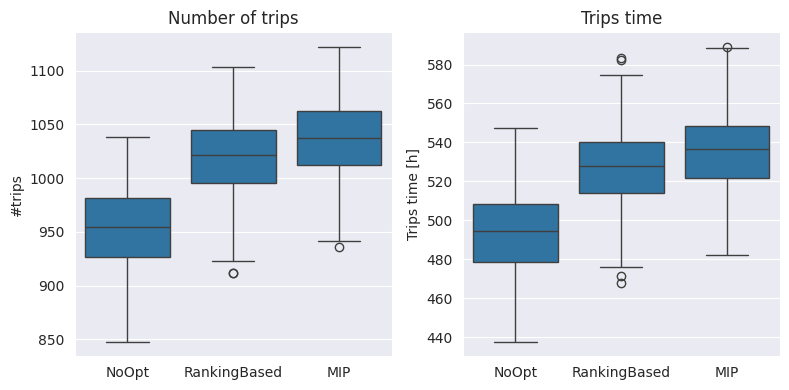}
    \caption{Distribution of trips count and time}
    \label{fig:boxplot-ranking_based_mip}
\end{figure}

While the solution utilizing local MIP optimization appears to outperform ranking-based algorithms, it does present certain disadvantages. First, tuning this method proved to be quite challenging. Both solutions utilize the same hyperparameters: relocation threshold, demand weight, and trip time weight. In contrast to the greedy algorithm, which is effective across a broad range of hyperparameters and straightforward to adjust, the local MIP optimization solution was notably demanding to tune. In total, we ran about 2000 optuna trials to select acceptable hiperparameters, and the result was rather accidental. Considering the average execution time of the scenario, the total tuning process took about 7 hours. Second, the identified hyperparameters are not yet optimal, as evident in Table~\ref{tab:rb_vs_local_mip} where relocations occur 2.5 times more often than scooter transits. 
As mentioned in Section\ref{subsec:model_calibration} in such case the introduced greedy ranking based algorithm also achieved superior scores, reaching improvements of up to 14\%. This disproportion is unsuitable for a real-world deployment, as it fundamentally transforms the business model from a car-sharing service to a car-hailing one.

\subsubsection{Scalability}
\label{subsec:scalability}

The aim of the scalability test was to analyze the algorithm’s execution time depending on the number of zones and the staff size. We assumed that the algorithm’s computational complexity can be approximated as $\mathcal{O}(kN)$, where $k$ denotes the size of the relocation crew and $N$ the number of service zones.
The procedures \textsc{schedule\_relocation} (see Algorithm~\ref{alg:schedule-relocation}) and \textsc{schedule\_transits} consist of double loops; however, the control flow bypasses zones lacking staff presence, resulting in the outer loop being executed no more than $k$ times.

The experiment was conducted using the following setup: the service area of Kraków was divided into $N=365$ hexagonal zones with a side length of 500 meters. Identical grid size is used in Traficar's internal demand heatmaps. Travel demand and vehicle presence data were aggregated into matrices with a temporal resolution of 15-minute intervals, yielding matrices of size $N \times N \times T$, where $T=96$.

Next, we prepared 266 different configurations that varied according to the number of zones $N_p$ ranging from 5 to 365, and the size of the staff $k_p$ between 3 and 20.
For each configuration, a random subset of zones was selected, forming a subproblem $P(N_P,k_P)$. For each subproblem 100 scenarios comprising 96 time steps were executed. The scenario matrices were sliced accordingly. Vehicle counts were redistributed across the selected zones. Algorithm execution times (excluding simulation steps) were recorded for each scenario and mean values and standard deviations were calculated. The baseline predictors were used: the predicted vehicle count was set as $\hat{x}_v[t+h] = x_v[h]$, and demand forecasts were taken directly from historical tables. However, inference with a simple linear regressor was simulated. 
The calculations were performed on a Ray platform configured to use 12 threads and took approximately 30 minutes.

Results of experiments are displayed in Figure~\ref{fig:scalability}. It can be noticed, that even for the largest instances, scenario execution times remained below 0.5 seconds. Surprisingly, the empirical time complexity turned out to be rather $\mathcal{O}(N_P^2)$, whereas the staff size $k_P$ had relatively small impact on execution time. This phenomenon can be explained by the fact that in many cases relocation and scooter transit times take longer than one time interval, hence at a given time $t$ many staff members are engaged and thus excluded from relocation decision-making.
In all configurations tested, the algorithm consistently outperformed random allocation. However, imbalances between relocations and scooter transfers were observed, indicating the need for calibration for specific deployments. This also explains the observed anomaly for the size of the staff $k_P=20$ and the number of zones greater than 300. In all these cases, the staff was mostly engaged in car travels.

\begin{figure}[H]
{
  \centering
  \includegraphics[width=0.5\linewidth]{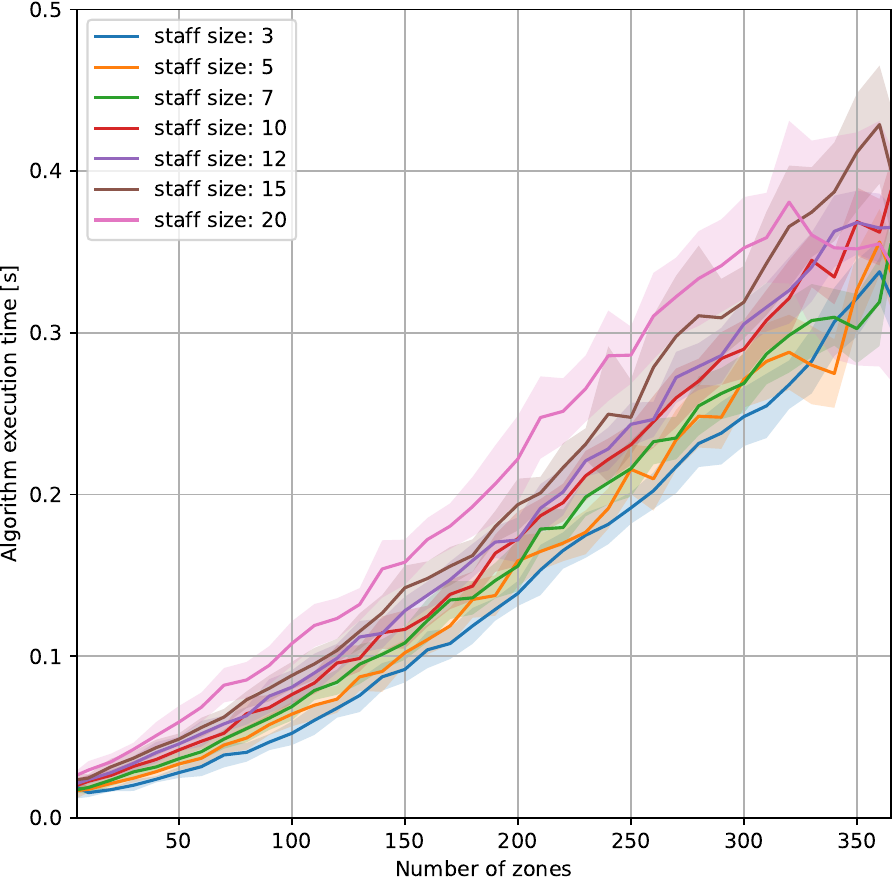}
  
  \caption{Algorithm execution time (simulation excluded, T = 96 time steps) versus number of zones, across different staff sizes. Results are averaged over 100 scenarios per configuration; shaded areas represent standard deviations.}
  \label{fig:scalability}
}
\end{figure}

\subsubsection{Discussion}

Results of benchmarking tests discussed in Section~\ref{subsec:impact-zoning} revealed very strong impact of clustering  method on performance of vehicle relocation algorithm. This is due to the fact, that partitioning into zones ignoring temporal patterns of demand and supply, may aggregate within clusters locations that compensate these factors. 

The algorithms for determining the locations of stations or charging stations described in Section~\ref{subsec:review-zoning-clustering} also indirectly create zones of influence for these facilities; however, their objectives are completely different from those of our zoning algorithm. They tend to be more interested in locations where long-term demand is expected to be high, while the algorithm being proposed emphasizes temporal trends across all levels of demand, as well as road proximity of locations within established zones.

At this point, it is worth also comparing our approach with that described in \citep{weikl20152}. The authors proposed a solution consisting in dividing a CS service area in Munich, Germany into 15 large zones. Vehicle relocations occur in two steps, first between large zones and then inside them  taking into account much smaller areas forming a hexagonal grid. The relocation algorithm seeks to determine the optimal number of vehicles in the zones based on historical data and aims to achieve this state by relocating vehicles at the lowest possible cost. A MIP model is proposed, which is solved using CPLEX software within a short time. In the next step a rule based algorithm is applied which directs to drivers either to charging stations or hot-spots. 

In our approach, we segment the service area into considerably smaller zones. Our objective is not to achieve an optimal supply level based on historical data; instead, we focus on foreseeing imbalances and making relocation choices accordingly. Consequently, our strategy is designed to be mode dynamic, aiming to address supply and demand variations over time. Our deployment concept also incorporates a two-phase relocation process. As drivers near their target zone, they are expected to utilize a hotspot map (available within the system) to select an optimal parking location. The zones we establish are quite small, encompassing areas accessible within a brief driving range, to facilitate parking within the designated area after a relatively short search.

The experiments described in Section~\ref{subsec:benchmarking-predictors} showed that the algorithm is not sensitive to the type of predictor used. It should be noted, however, that for the adopted zonal division, all predictors employed—even the simplest ones—demonstrated good performance. Their mean coefficients of determination (R\textsuperscript{2}) ranged from 0.89 to 0.94. However, in the case of the weakest predictor, xgboost, its negative impact on the results was noticeable.

The local optimization approach using MIP, described in Section~\ref{subsec:local-optimization}, yielded very promising results in terms of increasing travel time and the number of trips. However, the difficulties encountered with tuning, and in particular persisting imbalance between number of relocations and transits, make it a less attractive option for deployment.

The proposed algorithm coupled with simulation environment scales well computationally. 
Potential limitations are not due to the algorithm itself but rather to the scarcity of data. During the setup, historical data are preprocessed and space-time grids are filled, e.g., $N\times T$ for user activity and $N\times N \times T$ for travel times between zones and distribution of travels. With growing $N$ these grids become more and more sparse. 

Table~\ref{tab:table-densities} shows the matrix densities obtained for two examples of years 2022 and 2024. For 63 zones, one year of operational data from TraffiCar company (e.g., from 2024) results in 23.65\% matrix fill rates for travel time or frequency matrices. Missing values can often be imputed using smoothing or borrowing from adjacent time periods. For the 365-zone hexagonal grid, the fill rates were match worse: 1.53\% (2022) or 1.94\% (2024). 

Although travel frequency matrices are used only for simulation, travel times are essential for relocation ranking. The core steps of the algorithm include the computation of the imbalance factor $U$ for each zone according to formula (\ref{eq:imbalance}) and then two ranking scores $R_r$ and $R_t$ using formulas (\ref{eq:ranking_rel}) and (\ref{eq:ranking_tr}). The values $U$ are calculated from the output of the predictors. These computations are basically very fast for linear models, or can be cached in case of more complex algorithms. We decided to use 15-minutes sampling time, which seems appropriate for this kind of data.

However, obtaining realistic travel time values may constitute an issue in the case of scarcity of historical data. For each idle crew member, the travel time values for all other $N$ zones must be estimated. If historical data is unavailable, the travel time can be approximated based on distance in a road network, e.g. using pgRouting library, or by querying an external service like Google Maps API. The first solution is feasible and quite fast. The latter approach might be computationally expensive, e.g. assuming optimistically query time equal to 200~ms, and $N=365$ zones the total querying time for one employee would be equal to 73 seconds. However, the number of such queries can be significantly reduced by a simple heuristic that discards distant zones.

\begin{table}[H]
{
    \centering
    \begin{tabular}{|c|cc|cc|}
         \hline
         Year &  \multicolumn{2}{|c|}{$N=63$ (zones)}   & \multicolumn{2}{|c|}{$N=365$ (hex grid)}\\
         \hline
             &trips&act  &trips  &act \\
         \hline
         2022&21.52\%&99.42\%&1.53\%&94.13\% \\
         2024&23.65\%&99.52\%&1.94\%&92.97\% \\
         \hline
    \end{tabular}
    \caption{Densities of matrices for space-time grids}
    \label{tab:table-densities}
}
\end{table}

Another problem that arises is \emph{concept drift}: estimates of travel times are based on historical data, which should be collected over a sufficiently long period. We selected one year as the basic interval. We have chosen a year as the standard time frame. However, significant alterations in the road network, such as the introduction of new peripheral routes linking distant areas or the closure of a bridge for renovation, can render these data obsolete.

In summary, the main potential limitations for deployment include:

\begin{itemize}
  \item the availability and completeness of historical travel time data, or the need to compute these externally;
  \item prediction quality, which improves with spatial or temporal aggregation and sufficient data density;
  \item potential bottlenecks, when estimating travel times for multiple zones in real time;
  \item concept drift and adequacy of collected historical data.
\end{itemize}

\section{Conclusions}
\label{sec:conclusions}

The article introduces a two-part solution addressing the vehicle relocation problem in a free-floating car-sharing service. The first part involves an algorithm that segments the service area into several dozen zones. The dimensions and count of these zones are designed to facilitate the use of discrete optimization algorithms, ensuring the zones have similar traits, such as proximity along road networks, temporal behavior patterns, and road density. This zoning process employs an agglomerative clustering algorithm and is validated by its ability to predict vehicle availability numbers accurately. The second part consists of a newly designed algorithm for orchestrating vehicle relocations and crew movements. This algorithm relies on forecasts of vehicle availability and demand within the zones, making decisions based on ranked lists that factor in these predictions. It has been fine-tuned and tested by comparing its outcomes to random simulations and exact algorithms within a Mixed Integer Programming (MIP) model. Furthermore, the article presents an analysis on the effectiveness of the algorithm relative to the size of the crew,  results of the scalability experiment, predictors benchmarking, comparison with a hybrid solution using local optimization of MIP model, and discussion of technical issues related to prospective deployment.

When comparing the upper bound attained with the Gurobi solver for the MIPS model, the algorithm we proposed delivered unexpectedly favorable outcomes: 8.44\% as opposed to 19.6\% for identical parameters (however, with the stipulation that decisions in the MIPS model also pertained to trip selection, which we reject as not aligned with the business rules). We had anticipated an increase of about 3\%-4\% instead.

It is important to note that the research utilized actual data from a major car-sharing provider in Poland. The solution is tailored to match the platform's specific conditions and demands, with a particular focus on business conditions, the practicality of prediction algorithms for collected data, and a strong emphasis on real-time decision-making.

As the car-sharing platform progresses and introduces new operational modes and features, a future research focus will be the development of more sophisticated models and specialized algorithms to address emerging challenges. This encompasses novel platform functionalities, such as user-driven relocation, where users receive incentives for returning vehicles to high-demand areas, dynamic pricing based on local demand, or the capability to schedule vehicle delivery to a specified location and time. 

Another avenue of research could involve the use of various optimization techniques to guide decisions about staff actions, such as vehicle relocation, scooter movement, and other strategies in advanced models. Potential algorithms for this purpose may incorporate reinforcement learning or population-based metaheuristics applied at specific time points, using a surrogate objective function.

Finally, demand prediction for free-floating car-sharing systems appears to be a challenging problem in the context of fine-grained space-time models. In our work, we used basic forecasting algorithms, which can be upgraded to more sophisticated models, e.g. Informer networks or Graph Neural Networks (GNNs).

\bibliography{cas-refs}

\end{document}